\documentclass[conference]{IEEEtran}
\IEEEoverridecommandlockouts

\usepackage{enumitem}
\newtheorem{theorem}{Theorem}
\usepackage{hyperref}
\usepackage{subcaption}
\usepackage{graphicx}
\usepackage[version=4]{mhchem}
\usepackage{siunitx}
\usepackage{algorithm}
\usepackage{mathrsfs}
\usepackage{cite}
\usepackage{amsmath,amssymb,amsfonts,bm}
\usepackage{textcomp}
\usepackage{xcolor}
\usepackage{longtable,tabularx}
\usepackage[bottom= 54pt, right= 54pt, left= 54pt, top= 54pt]{geometry}
\usepackage[noend]{algpseudocode}
\begin{document}

\title{\vspace*{18pt}Model-Free Source Seeking by a Novel Single-Integrator with Attenuating Oscillations and Better Convergence Rate: Robotic Experiments}


\author{Shivam Bajpai, Ahmed A. Elgohary, and Sameh A. Eisa, \IEEEmembership{IEEE member}

\thanks{Shivam Bajpai is a master's student in the Department of Aerospace Engineering and Engineering Mechanics, University of Cincinnati, Ohio, USA (e-mail: bajpaism@mail.uc.edu). }
\thanks{Ahmed A. Elgohary is a PhD student in 
the Department of Aerospace Engineering and Engineering Mechanics, University of Cincinnati, Ohio, USA (e-mail: elgohaam@mail.uc.edu).}
\thanks{Sameh A. Eisa is an assistant professor in the Department of Aerospace Engineering and Engineering Mechanics, University of Cincinnati, Ohio, USA (e-mail: eisash@ucmail.uc.edu).}
}

\maketitle
\begin{abstract} 
In this paper we validate, including experimentally, the effectiveness of a recent theoretical developments made by our group on control-affine Extremum Seeking Control (ESC) systems. In particular, our validation is concerned with the problem of source seeking by a mobile robot to the unknown source of a scalar signal (e.g., light). Our recent theoretical results made it possible to estimate the gradient of the unknown objective function (i.e., the scalar signal) incorporated in the ESC and use such information to apply an adaptation law which attenuates the oscillations of the ESC system while converging to the extremum (i.e., source). Based on our previous results, we propose here an amended design of the simple single-integrator control-affine structure known in ESC literature and show that it can functions effectively to achieve a model-free, real-time source seeking of light with attenuated oscillations using only local measurements of the light intensity.      
Results imply that the proposed design has significant potential as it also demonstrated much better convergence rate. We hope this paper encourages expansion of the proposed design in other fields, problems and experiments. 
    
\end{abstract}

\section{Introduction}
\begin{figure}[h]
    \centering
    \includegraphics[width=0.45\textwidth]{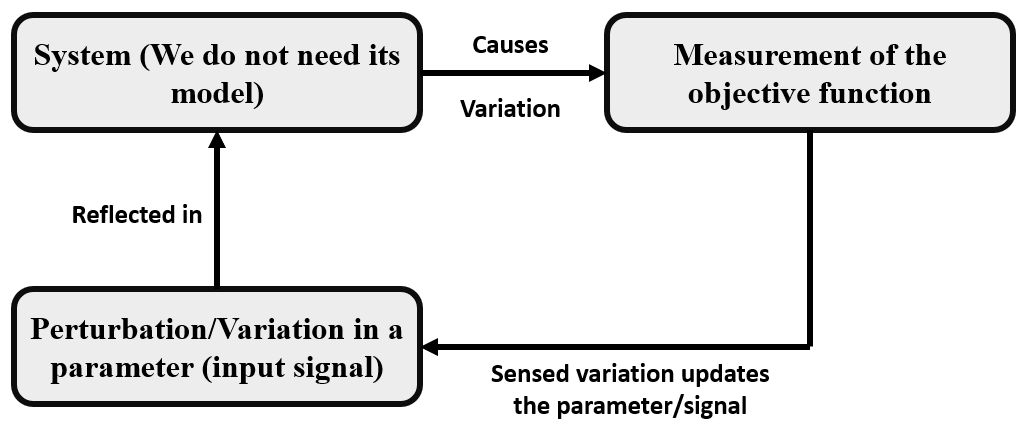}
    \caption{Simple diagram explaining ESC systems.}
    \label{fig:simple_diagram} 
\end{figure}

Extremum seeking control (ESC) systems \cite{tan2010extremum} are model-free, real-time adaptive control methods which aim at steering a given dynamical system to the extremum (maximum/minimum) of an objective function \cite{ariyur2003real} that may not be known expression-wise. Moreover, the way ESC systems operate make them very desirable to solve many problems in many fields as they only require a perturbation action and measurements of the objective function corresponding to the actuated perturbations \cite{ariyur2003real,scheinker2017model}. 
Via feedback of the objective function measurements, the ESC system updates the parameter or the control input to drive the system towards the extremum point. A simple diagram summarizing this idea is provided in figure \ref{fig:simple_diagram}.

\begin{figure}[h]
    \centering
    \includegraphics[width=0.35\textwidth]{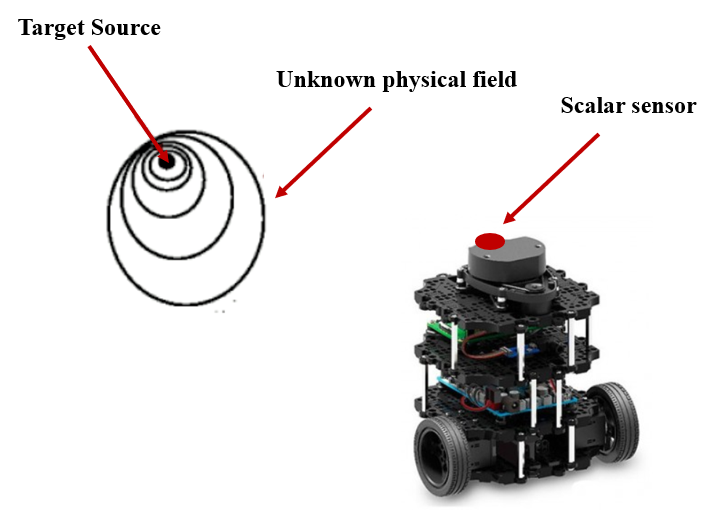}
    \caption{Sensor-based source seeking}
    \label{fig:Target_Source}
\end{figure}

Broadly speaking, two types of ESC systems can be found in literature: (i) ESC designs based on the classic structure \cite{krstic2000stability,krstic2000performance}; and (ii) ESC systems that are in control-affine forms \cite{durr2013lie,grushkovskaya2018class}. Both kinds can be studied/analyzed using averaging tools \cite{pokhrel2023higher}. ESC systems have been applied in unmanned systems, robotics, multi-agent systems, and bio-mimicry (e.g., \cite{ariyur2003real,durr2013lie,labar2018gradient,eisa2023, abdelgalil2023singularly,pokhrel2024extremum,zhuo2017source,cochran2009source,cochran2008gps,moidel2024reintroducing,pokhrel2022novel}). Many of the ESC applications in the mentioned fields fall under what is known as the ``source seeking" problem. That is, the ESC system steers a dynamical system (e.g., robot) towards the source point of a signal given that this source represents the maximum/minimum  of intensity or strength of said signal. Hence, the ESC advantage in these kinds of problems (i.e., source seeking) is that it will not require the mathematical expression of the signal distribution in the spacial domain or any global information for that manner (e.g., GPS). In fact, the ESC source seeking problem is solvable via the local measurement of the signal using, for example, sensors. Hence, ESC systems enable model-free, real-time source seeking. Figure \ref{fig:Target_Source} provides the idea of source seeking. However, when it comes to experimental efforts, verification, and deployments, most efforts in literature utilize the classic ESC structure or designs based on it, for the source seeking problem (e.g., \cite{ghods2011extremum, xu2022fast, bulgur2018light, oliveira2014monitoring}). In literature, it is quite rare to find experimental deployments or efforts involving control-affine ESC structures in general or using them in source seeking; the only work we were able to find \cite{grushkovskaya2018family} (to the best of our knowledge) which verifies experimentally the earlier theoretical results developed by the same authors in \cite{grushkovskaya2018class}.   

\textbf{Motivation.} Experiments of ESC using robotics, multi-agents, or unmanned systems can be challenging in general. This is due to the continuous oscillations issue in both the control input (e.g., the linear/angular velocity) and about the extremum (usually ESC systems stabilize in a limit cycle and practical stability sense \cite{krstic2000stability,durr2013lie,pokhrel2023higher}). The  continuous oscillations in the control input have been shown to be not much of an issue and definitely not an obstacle in real-world experiments of source seeking (e.g., \cite{ghods2011extremum, xu2022fast, bulgur2018light, oliveira2014monitoring}). In fact, we produced experiments using a robot (Turtlebot3) \cite{turtlebot3-emanual} for source seeking of light by classic ESC replicating the available literature of this problem -- see our YouTube video \cite{video_reference}. On the other hand, simple control-affine ESC structures can be beneficial to use in the source seeking problem given their overall simpler design, control laws, guaranteed stability \cite{durr2013lie,grushkovskaya2018class,pokhrel2023higher}, and the fact that many of them, like the single-integrator design (see \cite{durr2013lie}) have much smaller number of parameters to tune compared to many classic ESC designs. However, the authors in \cite{grushkovskaya2018family} stated in section IV.B that the control law ($\dot{x}=u=$$f(x)\sqrt{\omega} u_1+\sqrt{\omega} u_2$), which is the basis of the simple single-integrator design, performed the ``worst" in their experiments.
In fact, the same authors showed theoretically in \cite{grushkovskaya2018class} that single-integrator designs with basic quadratic function and sinusoidal control inputs will not achieve asymptotic convergence to the extremum, i.e., they will be persistently oscillating about the extremum. 

\textbf{Contribution.} In this paper, we take advantage from our recent theoretical developments in \cite{pokhrel2023control} which made it possible for control-affine ESC systems, including, simple single-integrator designs, to have attenuating oscillations as the system approaches the exteemum point (i.e., asymptotically convergent to the source point in a source seeking problem). In the mentioned work \cite{pokhrel2023control}, it was shown that structures eminent to have persistent oscillation per \cite{grushkovskaya2018class,grushkovskaya2018family} can in fact, by \textit{design}, be made structures with attenuating oscillations using what we call Geometric-based Extended Kalman Filer (GEKF) and an adaptation law which attenuates the control input signals in a stable and adaptive manner so the system perturbs as needed (i.e., the closer the system is from the extremum/source, the smaller perturbation it needs). We aim at using this development to introduce a \textit{novel single-integrator source seeking design with attenuating oscillations}. We provide both simulation results, but more importantly, novel experimental results for source seeking of light. The results validates our recent theoretical developments in \cite{pokhrel2023control} and show that one can achieve a model-free, real-time single-integrator-like source seeking with attenuating oscillations. In fact, our experiments demonstrate that the proposed design possesses \textit{remarkable better convergence rate} highlighting the potential of our work to be used in other control-affine ESC applications, problems and experiments.      

\section{Control-Affine Extremum Seeking with Attenuating Oscillations using Geometric-Based Kalman Filtering}
\begin{figure}[ht]
\centering
\includegraphics[width=0.45\textwidth]{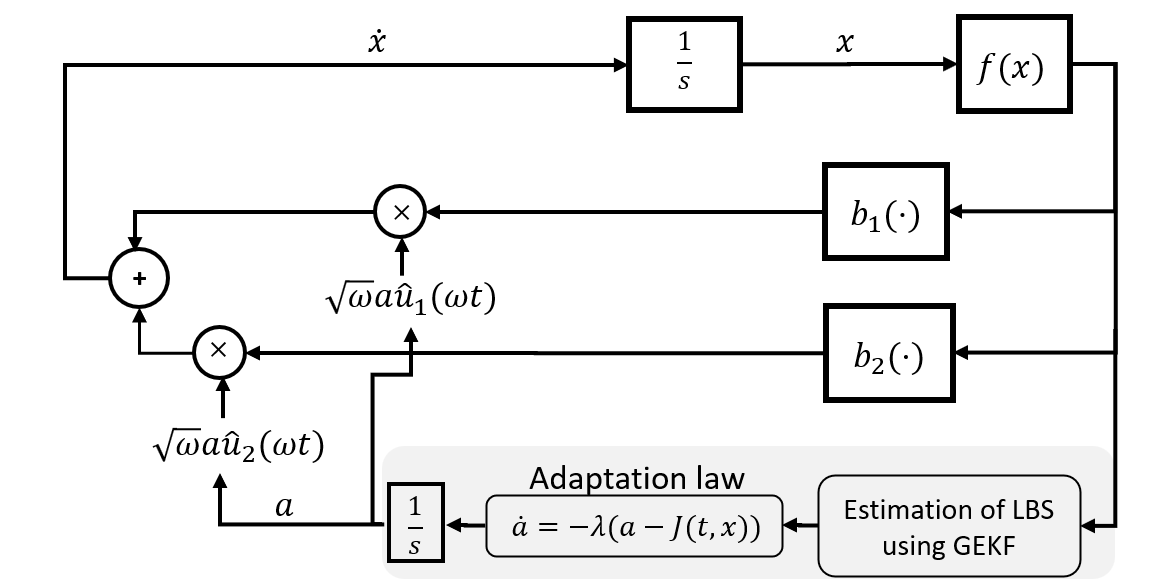}
\caption{The proposed ESC design in \cite{pokhrel2023control}.}
\label{fig:ESC_scheme}
\end{figure}

In this section, we briefly provide the background and preliminaries from our theoretical results in \cite{pokhrel2023control} which will be the basis for the main results provided in the next section. Control-affine ESC systems can be characterized as \cite{durr2013lie,pokhrel2023control}:
\begin{equation}\label{eqn:ESC_back}
    \dot{\bm{x}}=\bm{b_d}(t,\bm{x})+ \sum\limits_{i=1}^{m} \bm{b_i}(t,\bm{x})\sqrt{\omega}u_i(t,\omega t),\\
\end{equation}
with $\bm{x}(t_0)=\bm{x_0}\in \mathbb{R}^n$ and $\omega \in (0,\infty)$, where $\bm{x}$ is the state space vector, $\bm{b}_d$ is the drift vector field, $u_i$ are the control inputs, $m$ is the number of control inputs, while the vectors $\bm{b}_i$ correspond to the control vector fields. The control-affine ESC system in \eqref{eqn:ESC_back} can be approximated and characterized by what is known as the Lie bracket system (LBS) \cite{durr2013lie,pokhrel2023control}:
\begin{equation}\label{eqn:Lie_back}
    \dot{\bm{z}}=\bm{b_d}(t,\bm{z})+ \sum_{\substack{i=1\\j=i+1}}^m [\bm{b_i},\bm{\bm{b_j}}](t,\bm{z})\nu_{j,i}(t),
\end{equation}
with $\nu_{j,i}(t)=\frac{1}{T}\int_0^T u_j(t,\theta)\int_0^\theta u_i(t,\tau)d\tau d\theta$. The operation $[\cdot,\cdot]$ donates Lie bracket operation applied to two vector fields $\bm{b_i},\bm{b_j}: \mathbb{R} \times\mathbb{R}^n \rightarrow \mathbb{R}^n$ with $\bm{b_i}(t,\cdot),\bm{b_j}(t,\cdot)$ being continuously differentiable, and is defined as $[\bm{b_i},\bm{b_j}](t,\bm{x}):=\frac{\partial \bm{b_j}(t,\bm{x})}{\partial \bm{x}}\bm{b_i}(t,x)-\frac{\partial \bm{b_i}(t,\bm{x})}{\partial \bm{x}}\bm{b_j}(t,\bm{x})$. LBSs in \eqref{eqn:Lie_back} are gradient-like systems \cite{durr2013lie,grushkovskaya2018class,pokhrel2023control} that average \cite{pokhrel2023higher} the ESC system in  \eqref{eqn:ESC_back}. In \cite{grushkovskaya2018class}, a class of the control-affine ESC based on \eqref{eqn:ESC_back} was proposed to generalize many control-affine ESC systems in the literature (see \cite{grushkovskaya2018class,pokhrel2023control} for more details). Our work in \cite{pokhrel2023control} extended the mentioned generalized class and proposed a new ESC class which we will use in this paper:
\begin{align}
    \dot{\bm{x}}&=\sum \limits_{i=1}^n \left( b_{1i}(f({x}))\sqrt{\omega} a_i(t)\hat{u}_{1i}+ {b_{2i}}(f({x}))\sqrt{\omega} a_i(t) \hat{u}_{2i}\right) e_i \label{eqn:generalizedSystem},\\
    \dot{\bm{a}}&=\sum \limits_{i=1}^n \left(-\lambda_i ({a}_i(t)-{J}_i(t,\bm{x})) \right ) e_i,\label{eqn:law}
\end{align}
where $e_i$ denotes the $i^{th}$ unit vector in $\mathbb{R}^n$. Moreover, ${b}_{1i}$ and ${b}_{2i}$ are the vector fields associated with the control inputs $u_{1i}=a_i\sqrt{\omega}\hat{u}_{1i}(\omega t),u_{2i}=a_i \sqrt{\omega}\hat{u}_{2i}(\omega t)$, the objective function $f:  \mathbb{R}^n \to \mathbb{R}$, $a_i \in \mathbb{R}$ is the amplitude of the input signal, and $\lambda_i >0\in \mathbb{R}$ is a tuning parameter. Lastly, ${J}_i(t,\bm{x})$ represents the estimation of the right hand side of the LBS in \eqref{eqn:Lie_back} which can be estimated merely by estimation of the gradient of the objective function $\nabla f(\bm{x})$ as shown in our work \cite{pokhrel2023control}. Now, we impose the following assumptions:
\begin{enumerate}[label=A\arabic*.]
    \item 
    $b_{ji}, b_{ji}\in C^2: \mathbb{R} \to \mathbb{R}$, and for a compact set $\mathscr{C} \subseteq \mathbb{R}$, there exist $A_1, ..., A_3 \in [0,\infty)$ such that $|b_{j}(x)|\leq A_1,
    | \frac{\partial b_j(x)}{\partial x}|\leq A_2,
    |\frac{\partial [{b_j},{b_k}](x)}{\partial x}| \leq A_3$ for all $x\in \mathscr{C}, i={1,2};\; j={1,2};\; k={1,2}.$
    \item 
    $\hat{u}_{1i}, \hat{u}_{2i}: \mathbb{R} \times \mathbb{R} \to \mathbb{R} , i=1,2$, are measurable functions. Moreover, there exist constants $M_i \in (0,\infty) $ that $sup_{\omega t \in \mathbb{R}}|\hat{u}_i(\omega t)|\leq M_i$, and 
    $\hat{u}_i(\cdot)$ is T-periodic, i.e. $\hat{u}_i(\omega t + T)=\hat{u}_i(\omega t),$ and has zero average, i.e. $\int_0^T \hat{u}_i(\tau) d\tau = 0,$ with $T \in (0,\infty)$ for all $\omega t \in \mathbb{R}$.
    \item 
    There exists an ${x}^* \in \mathscr{C}$ such that $\nabla f({x}^*)=0, \nabla f({x})\ne 0$ for all ${x}\in \mathscr{C}\backslash \{{x}^*\}; f({x}^*)=f^* \in \mathbb{R}$ is an isolated extremum value. 
    \item Let the estimation error of ${J}_i(t,\bm{x})$ be ${\eta}_i(t)$, then ${\eta}_i(t): \mathbb{R} \rightarrow \mathbb{R},i=1,...,n$ is measurable function and there exist constants $\theta_0,\epsilon_0 \in (0,\infty)$ such that $|{\eta}_i(t_2)-{\eta}_i(t_1)| \le \theta_0|t_2-t_1|$ for all $t_1,t_2 \in \mathbb{R}$ and $\sup_{t \in \mathbb{R}}|{\eta}_i(t)| \le \epsilon_0$.
    Furthermore, $\lim_{t\to \infty} {\eta}_i (t)={0}$.
\end{enumerate}
Our ESC design \cite{pokhrel2023control} for $n=1$ is shown in figure \ref{fig:ESC_scheme}. We recall the following theorem from \cite{pokhrel2023control} which provides the stability of 
provides the stability of the ESC system \eqref{eqn:generalizedSystem}-\eqref{eqn:law}
\begin{theorem}\label{thm:proposed_theorem}
Let A1-A4 be satisfied with some $\omega \in (\omega^*,\infty), \omega^* >0$ and suppose $\exists \; t^* >t_0 =0$ such that $\forall t>t^*$  and $|J_i(t,\bm{z})| \le 1/t^p$ with some $p>1$ then (i) the equilibrium point $\hat{\bm{z}}^* \in \mathscr{C}$ is locally asymptotically stable for the estimated LBS of $\dot{\bm{z}}=J_i(t,\bm{z})$, (ii) $a_i$ in (\ref{eqn:law}) is asymptotically convergent to 0; and (iii) the system in (\ref{eqn:generalizedSystem}) is practically asymptotically stable.
\end{theorem}

\section{Main Results}
\begin{figure}[h]
      \centering
     \includegraphics[width=0.45\textwidth]{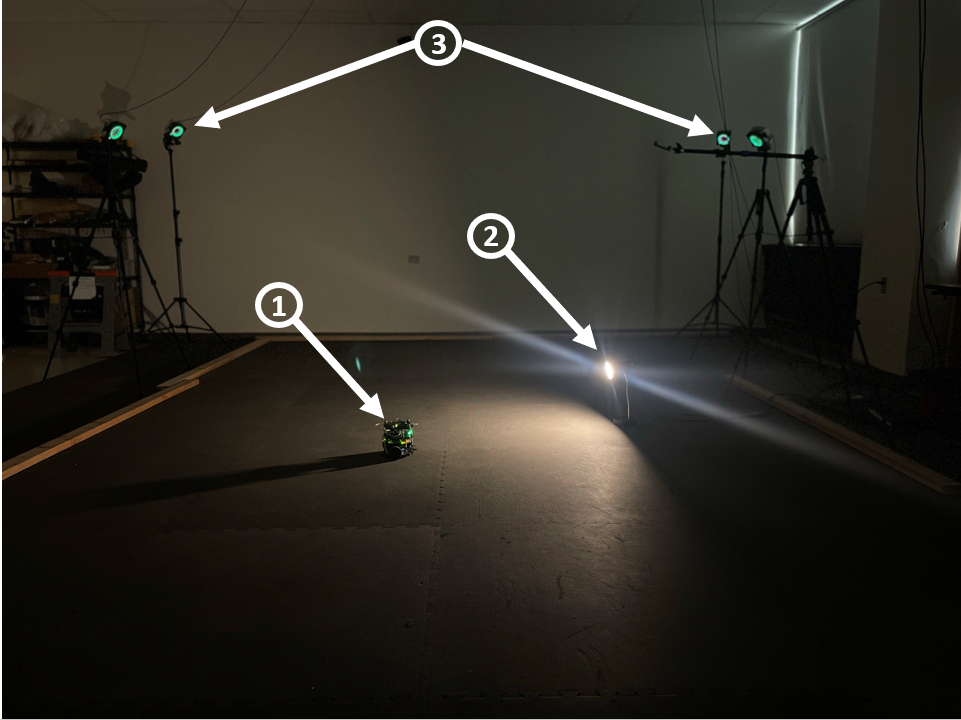}
     \caption{Overview of the experimental setup.}
     \label{fig:Setup}
  \end{figure}

For the experimental work utilized in this section, we use the Turtlebot3 (TB3) robot \cite{turtlebot3-emanual}. TB3 is a small, affordable, ROS-based, and differential drive mobile robot platform that is designed by ROBOTICS, for the purpose of research, education. It has two forms: burger, and waffle. In this work, we have used the burger TB3 type which has wheel base and wheel radius of 160mm and 33mm. In fact, the robot displayed in figure \ref{fig:Target_Source} is a TB3. An overview of the experimental setup in our lab is shown in figure \ref{fig:Setup}. There are three main components in the setup. First component (denoted by \#1) is the TB3. Second component  (denoted by \#2) is the light source utilized during the experiment, which can be replaced by any extremum/source in the context of ESC literature. Third component (denoted by \#3) is the motion capture system employed to track and visualize the planar trajectory of the TB3.


\begin{figure}[ht]
\label{fig:ESC_Design_with_EGKF}
    \centering
    \includegraphics[width=0.5\textwidth]{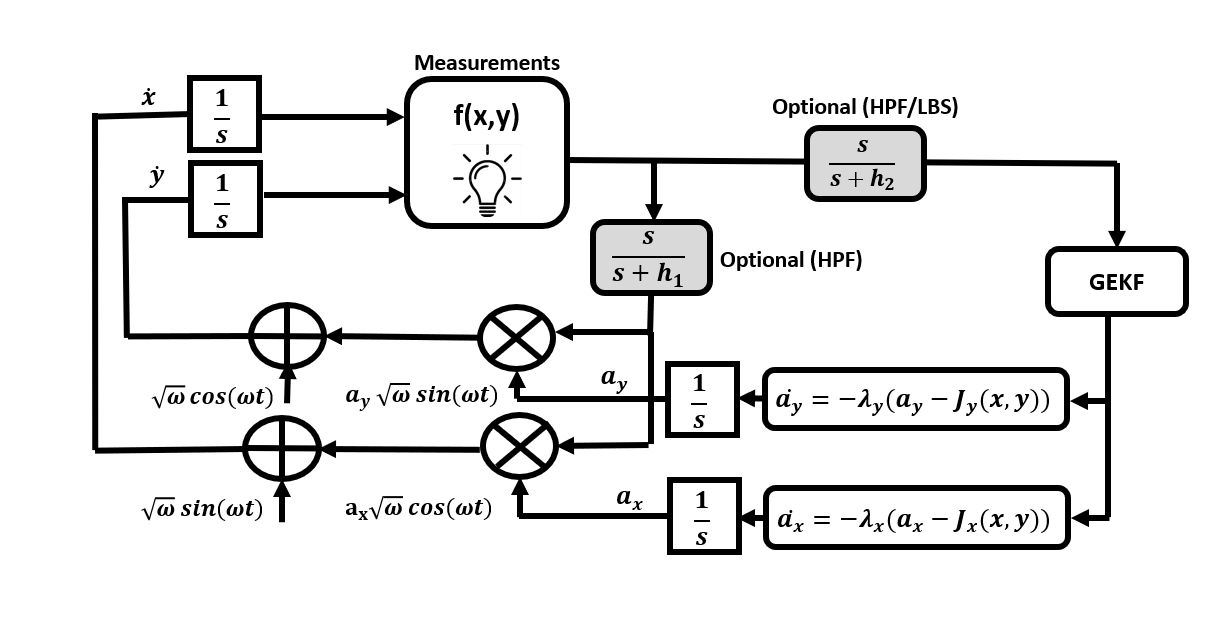}
    \caption{Proposed amended single-integrator design with GEKF and optional High/Low Pass Filters (HPF/LPS).}
    \label{fig:ESC Design with EGKF}
\end{figure}
\subsection{Novel Single-Integrator-Like Source Seeking Design}
Here we provide details on our proposed novel single-integrator-like source seeking design which stabilizes a system (e.g., robot like TB3) about the extremum/source with attenuated oscillations. Traditional single-integrator designs have been used in literature (e.g., \cite{durr2013lie}). In this work, we propose a novel (amended) single-integrator design based on our recent theoretical developments in \cite{pokhrel2023control} which is summarized in the previous section.  That is, we propose an amended single-integrator source seeking design with attenuating oscillations as shown in figure \ref{fig:ESC Design with EGKF}. Note that the proposed design in figure \ref{fig:ESC Design with EGKF} is slightly different from the theoretical results in \cite{pokhrel2023control} as we incorporate optional High/Low pass filters (HPF/LPF). The reason for introducing these optional filters into the design is to provide the user with the flexibility to attenuate undesired noise and select frequency ranges if needed. In our results, the inclusion of these filters has demonstrated the ability to enhance the transient performance of the system. Along the same lines of (\ref{eqn:generalizedSystem})-(\ref{eqn:law}), the proposed design which operates in a planar mode ($x$ and $y$ coordinates) can be represented as follows: 

\begin{equation}\label{eqn:eg3esc}
\begin{split}
    \dot{x}&= c \, f(x,y)\sqrt{\omega}  u_1(\omega t) + a_x \sqrt{\omega} u_2(\omega t),  \\
    \dot{y}&= -c \,f(x,y) \sqrt{\omega} u_2(\omega t) + a_y \sqrt{\omega} u_1(\omega t),\\
     \dot{{a}}_x&=-\lambda_x ({a_x}-{J_x}({x},{y})),\\
     \dot{{a}}_y&=-\lambda_y ({a_y}-{J_y}({x},{y})),
    \end{split}
\end{equation}
where $u_1(\omega t) = \sin(\omega t)$, $u_2(\omega t) = \cos(\omega t)$, $f(x,y)$ is the objective function, $\omega$ is the frequency, $c$ is a constant, $a_x$, $a_y$$ \in \mathbb{R}$ are the amplitude of the input signals for x and y states, respectively, $\lambda_x$, $\lambda_y$$ >0\in \mathbb{R}$ are tuning parameters, and $J_x (x,y)$ and $J_y (x,y)$ represent the estimation of the right-hand side of the LBS such that \cite{pokhrel2023control}:

\begin{equation}\label{eqn:Liebracket}
\begin{aligned}
J_x(x, y) &= \alpha_1 \nabla_{z_x} f(x, y) + \eta_1(t), \\
J_y(x, y) &= \alpha_2 \nabla_{z_y} f(x, y) + \eta_2(t),
\end{aligned}
\end{equation}
with $\alpha_1, \alpha_2$ are constants and $\eta_1, \eta_2$ are the estimation error of $J_x , J_y$, respectively.

As shown in \cite{pokhrel2023control}, the estimations $J_x$ and $J_y$ are good and viable for the validity of Theorem 1 only if the gradient estimation is accurate with vanishing error as $t \rightarrow \infty$. This is what we achieved in \cite{pokhrel2023control} via the novel filter we called Geometric-based Extended Kalman Filter (GEKF) \cite{pokhrel2023gradient}. 
For our design in this paper, as is the case in \cite{pokhrel2023control}, GEKF \cite{pokhrel2023gradient} works in real-time and is considered as a discrete-continuous extended Kalman filter that requires: 
\begin{enumerate}
    \item a measurement equation, which is the discrete part that relates the parameter we need to estimate (the gradient of the objective function $\nabla f(x,y)$) with the measurements we have access to (measurements of the objective function $f(x,y)$)
    \item the propagation model, which is the continuous part that governs the filer behavior as a continuous model between the measurements.
\end{enumerate} 
The GEKF procedure is summarized in Algorithm 1. However, the reader may refer \cite{pokhrel2023gradient} for more details on GEKF idea, steps, application and solved examples by detail. Also, the reader can refer \cite{pokhrel2024extremum} for an application using GEKF. In this paper, the GEKF states are formulated as follows:
\begin{align}
    \bar{\bm{X}}=\begin{bmatrix}
    \bar{x}_1\\
    \bar{x}_2\\
    \bar{x}_3\\
    \bar{x}_4\\
    \bar{x}_5
    \end{bmatrix}=\begin{bmatrix}
    \frac{K}{2} \nabla_{x} f(x,y)\vert _{t_1}\\
    \frac{K}{2} \nabla_{y} f(x,y)\vert _{t_1}\\
    \dot{\bar{x}}_1\\
     \dot{\bar{x}}_2\\
     f(x,y)\vert _{t_1}\\
    \end{bmatrix}_{5 \times 1},
\end{align}
where $\bar{\bm{x}}_1$ and $\bar{\bm{x}}_2$ are proportional to the gradient components we are estimating. Following the guidelines provided in \cite{pokhrel2023control,pokhrel2023gradient}, we take $k=(2/\sqrt{\omega}) \sin(\omega \Delta t/2)$ where $\Delta t$ is a short time step between the measurements. It is important to emphasize that the time step $\Delta t$ has to be significantly smaller than the periodic time. This requirement is due to the fact that both ESC systems and GEKF \cite{pokhrel2023control} perform more effectively at higher frequencies. Now, following a constant velocity propagation model similar to \cite{pokhrel2023control,pokhrel2023gradient}, we get:

\begin{align}\label{eqn:GEKFDynamics_eg3}
    \dot{\bar{\bm{X}}}=\begin{bmatrix}
    \bar{\bm{x}}_3\\
    \bar{\bm{x}}_4\\
    0\\
    0\\
    0\\
    \end{bmatrix}_{5 \times 1} + \bm{\Omega},
\end{align}
where $\bm{\Omega}$ represents the process noise as a random variable, $\bm{Q}$ is the covariance matrix for process noise (system noise) and the  Jacobian matrix $\bm{A}$ associated with the state dynamics is obtained as
\begin{align}\label{eqn:matrixA_case3}
    \bm{A}=\begin{bmatrix}
    0 & 0 & 1 & 0 & 0\\
    0 & 0 & 0 & 1 & 0\\
    0 & 0 & 0 & 0 & 0\\
    0 & 0 & 0 & 0 & 0\\
    0 & 0 & 0 & 0 & 0\\
    \end{bmatrix}_{5 \times 5}.
\end{align}

For the measurement updates, The Chen-Fliess series expansion is used as shown in detail in \cite{pokhrel2023control,pokhrel2023gradient} as follows: 

\begin{equation}\label{eqn:measurementeqn3}
\begin{aligned}
    f(\bm{x,y})\vert _{t_2} &= f(\bm{x,y})\vert _{t_1} + (\nabla f(\bm{x,y}) \cdot \mathbf{b}_1)\vert _{t_1} K\cos (\omega t)\\
    &\quad+ (\nabla f(\bm{x,y}) \cdot \mathbf{b}_2)\vert _{t_1} K \sin(\omega t) + {\nu}(t).\\ 
    \text{and} \\
    \mathbf{b}_1 &= \begin{bmatrix} c  \sqrt{\omega}f(x, y)& ; \quad
    a_x \sqrt{\omega}\end{bmatrix}, \\
    \mathbf{b}_2 &= \begin{bmatrix} -c \sqrt{\omega}f(x, y)&; \quad
    a_y \sqrt{\omega}\end{bmatrix},
\end{aligned}
\end{equation}
where $ t_2=t_1+\Delta t$. The residual terms follow a Gaussian distribution as measurement noise denoted as $\nu (t) \sim N(0,R)$, and R is the covariance matrix associated with noise measurement. Moreover, it is further assumed that the process noise and the measurement noise are uncorrelated. Now, we rewrite the measurement update equation as:
\begin{equation}\label{eqn:eg3meas1}
\begin{split}
    &f(\bm{x,y})\vert_{t_2} = \bm{h}(\bar{\bm{X}})+ {\nu} (t)\\
    &=f(\bm{x,y})\vert_{t_1} \\
    &\quad+ \Big( cf(\bm{x,y})\nabla_x f(\bm{x,y}) + a_x \nabla_y f(\bm{x,y}) \Big) K \cos (\omega t) \\
    &\quad+ \Big( a_y \nabla_x f(\bm{x,y}) - cf(\bm{x,y})\nabla_y f(\bm{x,y}) \Big) K \sin(\omega t) \\
    &\quad+ \nu(t) \\
    &= \bar{x}_5 + 2c\bar{x}_1\bar{x}_5 \cos (\omega t) + 2a\bar{x}_2 \cos (\omega t) \\
    &\quad+ 2a\bar{x}_1 \sin(\omega t) - 2c\bar{x}_2\bar{x}_5 \sin(\omega t) + \nu(t),
\end{split}
\end{equation}

where $t=(t_1+t_2)/2$. The Jacobian matrix associated with the measurement update equation is:
\begin{align}
    \bm{C}=\begin{bmatrix}
    2c\bar{x}_5 \cos(\omega t) + 2a_y \sin(\omega t)\\
    2a_x \cos (\omega t)-2c\bar{x}_5 \sin(\omega t)\\
    0\\
    0\\
    1+2c\bar{x}_1 \cos(\omega t)-2c\bar{x}_2 \sin(\omega t)
    \end{bmatrix}_{5 \times 1}
\end{align}
\begin{algorithm}
		\caption{Geometric-based Continuous-discrete Extended Kalman Filter} 
		\color{black}
		\label{alg:alg2}
			\begin{algorithmic}
			\State Initialize $\bar{\bm{X}}$ with some initial values.
			
			\State choose a sample rate as an output $T_{out}$ less than the sensors rates (used for measurments ).
			
			\State For each sample time $T_{out}$:\\
			for $i=1$ to $N$ do (Prediction Step)\\
			\quad $\bar{\bm{X}}=\bar{\bm{X}}+ (T_{out}/N) \dot{\bar{\bm{X}}}$\\
			\quad $\bm{P}=\bm{P}+\frac{T_{out}}{N}(\bm{AP+PA+Q})$,

			\State once the measurement is received $i$ then\\
			(Measurement step)\\
			\quad $\bm{C}=\frac{\partial \bm{h}}{\partial \bm{X}}(\bar{\bm{X}})$\\
			\quad $\bm{L} = \bm{PC}^T(\bm{R}+\bm{C} \bm{P C}^T)^{-1}$\\
			\quad $\bm{P}=(\bm{I}-\bm{L} \bm{C}) \bm{P}$\\
			\quad $\bar{\bm{X}}=\bar{\bm{X}}+\bm{L} (\bm{y}[n]-\bm{h}(\bar{\bm{X}}))$
			
			\State end if
		\end{algorithmic} 
\end{algorithm}
\begin{figure*}[h]
  \centering
  \begin{subfigure}{0.3\textwidth}
    \includegraphics[width=\linewidth]{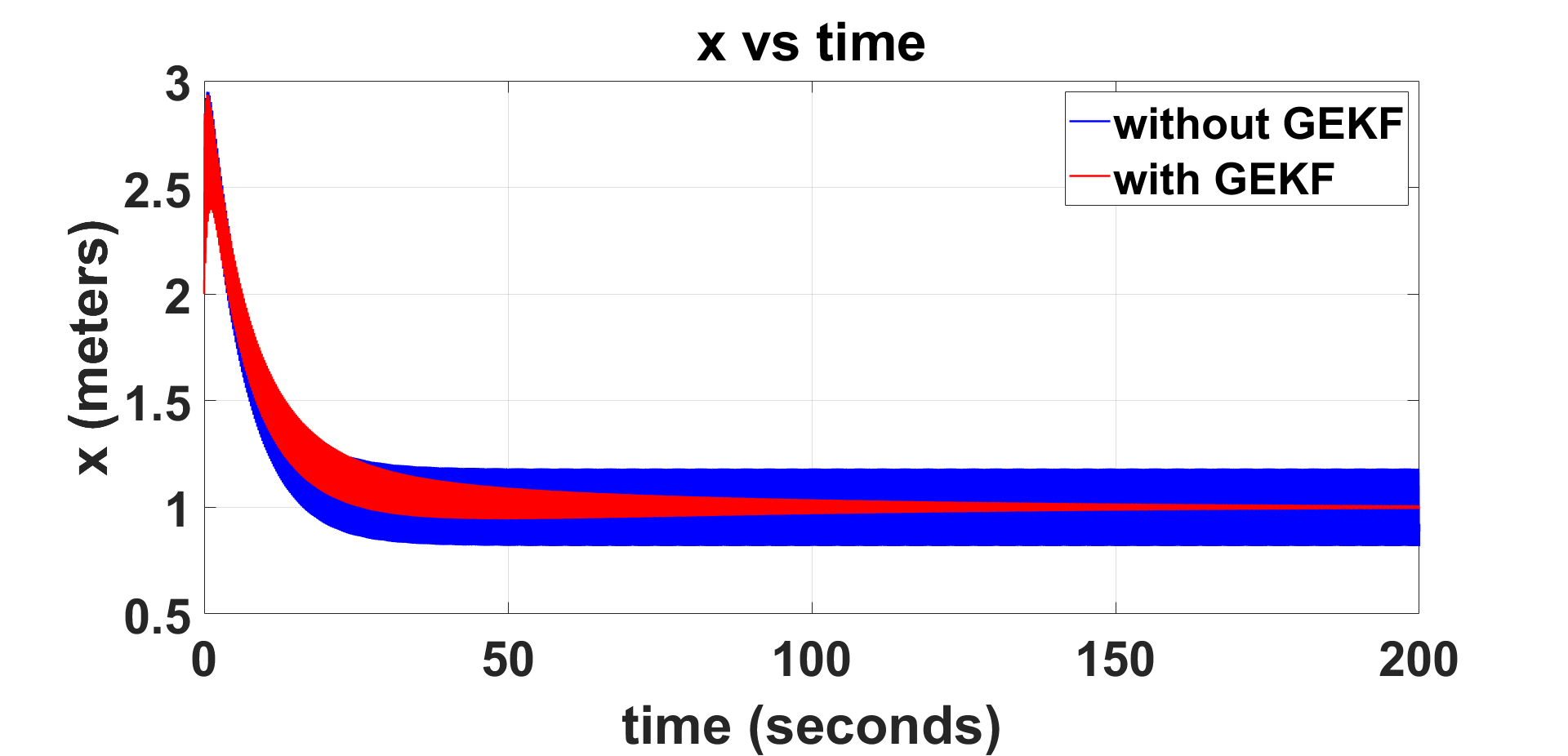}
    \caption{$x$ position vs. time.}
    \label{fig:X_Vehicle_1}
  \end{subfigure}
  \hfill
  \begin{subfigure}{0.3\textwidth}
    \includegraphics[width=\linewidth]{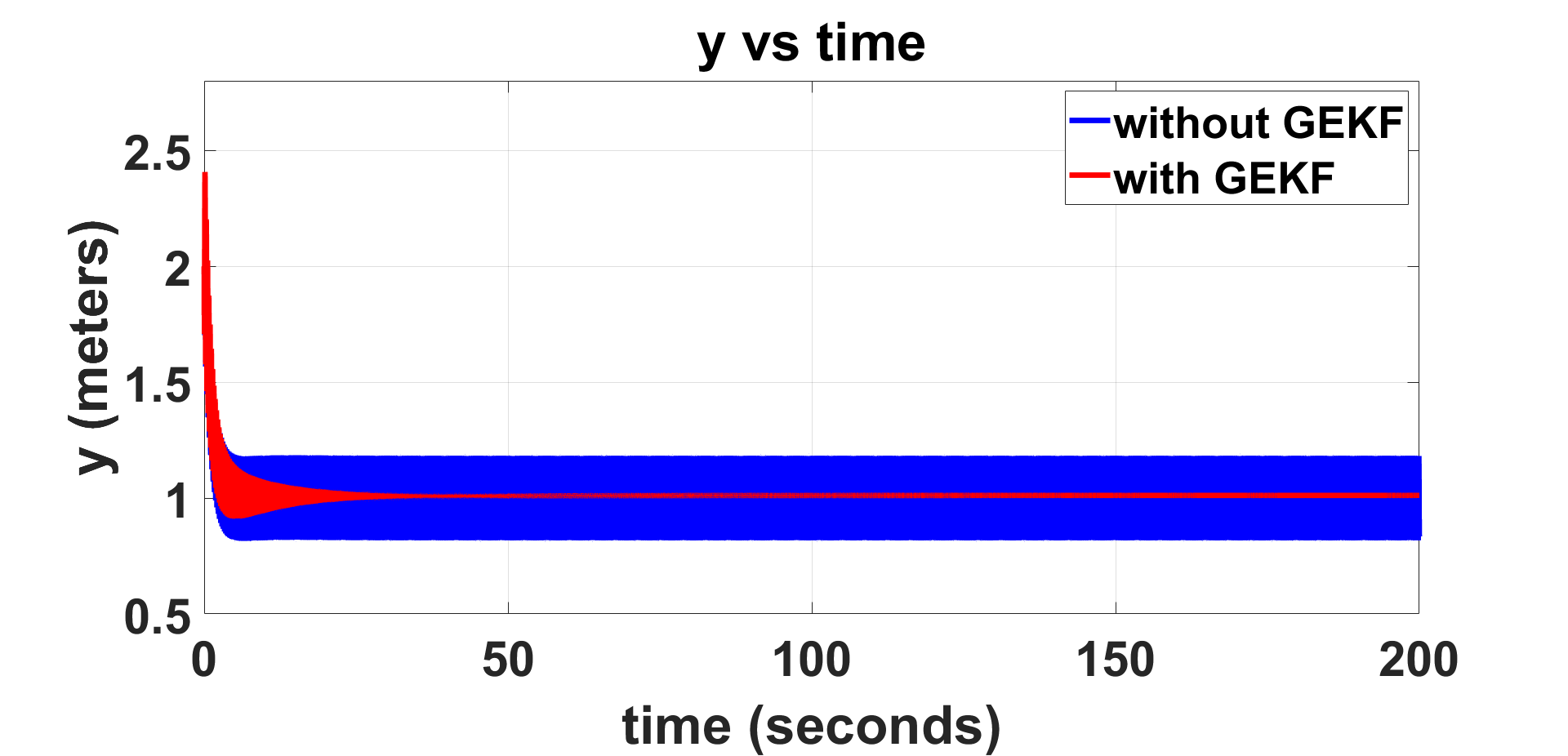}
    \caption{$y$ position vs. time.}
    \label{fig:Y_Vehicle_1}
  \end{subfigure}
  \hfill
  \begin{subfigure}{0.3\textwidth}
    \includegraphics[width=\linewidth]{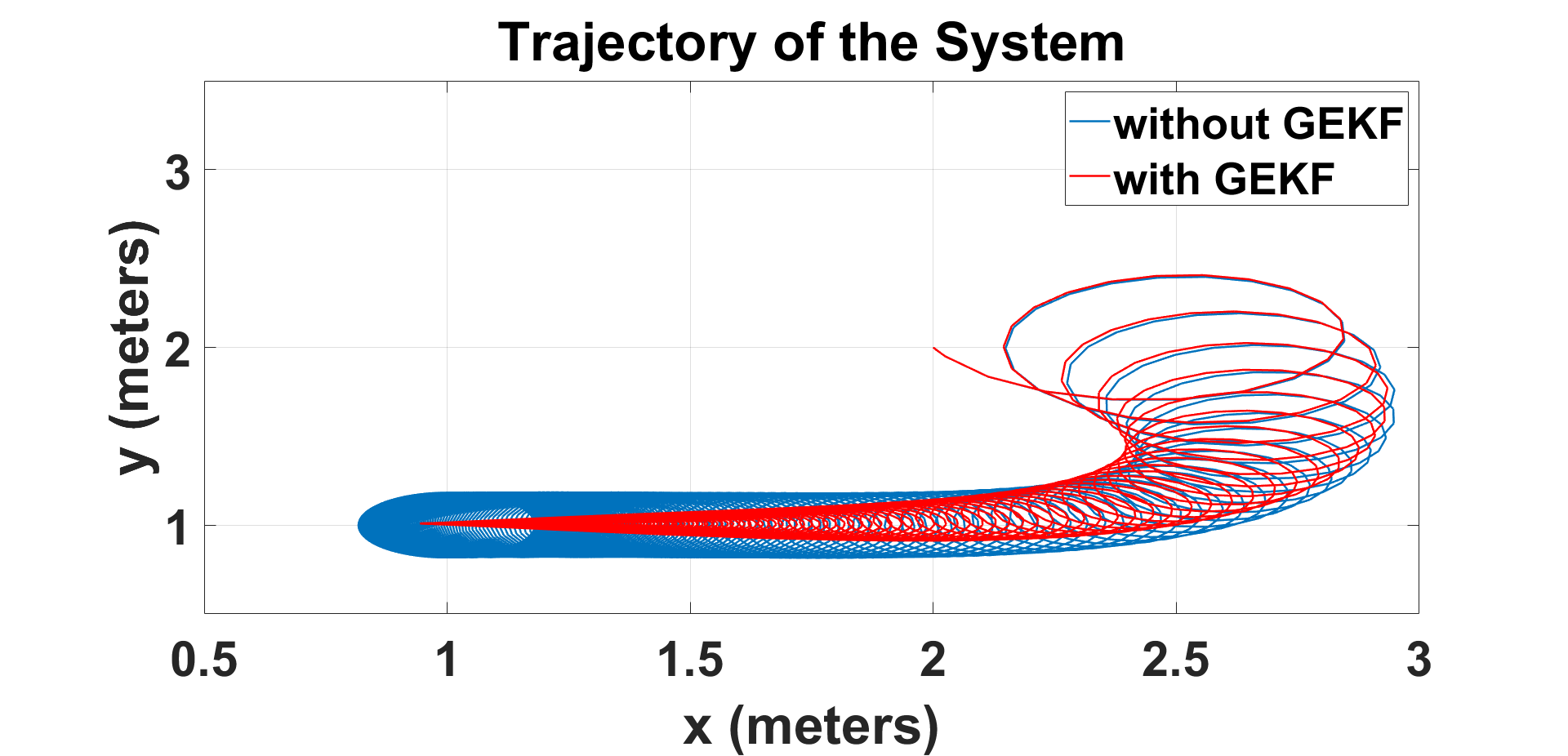}
    \caption{Planar plot of $x$ vs. $y$.}
    \label{fig:X_Y_V1}
  \end{subfigure}
    \caption{Simulation of $x$ and $y$ coordinates with obtaining measurements from a known objective function. Our proposed design (red) attenuate oscillations successfully when compared with literature (blue).}
  \label{fig:Mathmatical_Model_simulation}
\end{figure*}

With the measurement update equation \eqref{eqn:eg3meas1} in place and the propagation model \eqref{eqn:GEKFDynamics_eg3} in place, Algorithm 1 can be applied effectively as per \cite{pokhrel2023control,pokhrel2023gradient} and our proposed design in figure \ref{fig:ESC Design with EGKF} can be put in action. In the next subsections our proposed design will be verified by simulations and mostly real-time, real-world experiments. Here we layout our efforts in conducting said simulations and experiments in the next subsections. Whenever possible, we compare our proposed design in figure \ref{fig:ESC Design with EGKF} (amended single-integrator with GEKF and adaptation law for the attenuation of oscillations) vs. the traditionally found single-integrator design in literature (e.g., \cite{durr2013lie}). The first phase of our verification results is provided in subsection 3.B. In that phase, we take the objective function measurements from a known mathematical expression which we do not use in any analytical computations (e.g., computing $\nabla f(x,y)$). This enables us to test the design in a more controlled manner and environment. For instance, by taking the measurements from a mathematical expression of an objective function, we can: (i) run a simulation for the experiment before the hardware implementation; (ii) guarantee the applicability of assumptions A1-A3 in the real-world experiment, and (iii) have less noisy environment so that the performance of GEKF is more ideal in its first time ever experimental implementation.     
The second phase of our verification results is provided in subsection 3.C. In that phase, we take the experiments in a step forward where we use a light sensor for measuring light intensity, aiming at model-free, real-time light source seeking by the TB3. In this phase, we obviously are working in a complete model-free fashion in all fronts (no equations for TB3 model, sensor, or light intensity/distribution). Lastly, in subsection 3.D, we provide some comments on how Theorem 1 condition seems to have been observed in our experimentation.  

\subsection{Simulations and Experimentation of the Proposed Design with Known Objective Function}
As elaborated at the end of the previous subsection, here we provide details, results and observations of our first phase. In this phase, we implement our design with a know mathematical expression of the objective function $f(x,y)$ but only for obtaining measurements. We first present our results via simulations to test and verify the effectiveness of our design. We used MATLAB and Simulink\textsuperscript{\textregistered} in said simulations. Now, we provide details on the conducted simulations as follows. For the states $(x,y)$, the initial conditions are taken as (2,2). For the GEKF,  the initial value of the covariance matrix $P$ is taken as $[4,4,4,4,4]^T$, sample rate $T_{out} = 0.1$, $\bm{Q}=0.05\bm{I}$ with $\bm{I}$ being the identity matrix, $R=0.5$ and $N=10$. For the ESC parameters we take $\omega=30$, and $c=0.3$. For the adaptation law we take $a_x(0)=1$ and $a_y(0) = 1$; note that for simulations with traditional single-integrator (i.e., no GEKF and no adaptation law), $a_x=a_y=constant=1$, $\lambda_x=0.015$ and $\lambda_y=0.0995$. We used all optional high pass filters (HPFs) such that $h_1=h_2=1$. The objective function we used for obtaining the measurements is $f(x,y)=10-\frac{1}{2}(x-1)^2- \frac{3}{2}(y-1)^2$ which clearly yields an extremum (maximum) at $f^*=10$ when $(x,y)=(1,1)$.

\begin{figure}[ht]
  \centering
  \begin{subfigure}{0.45\textwidth}
    \includegraphics[width=\textwidth]{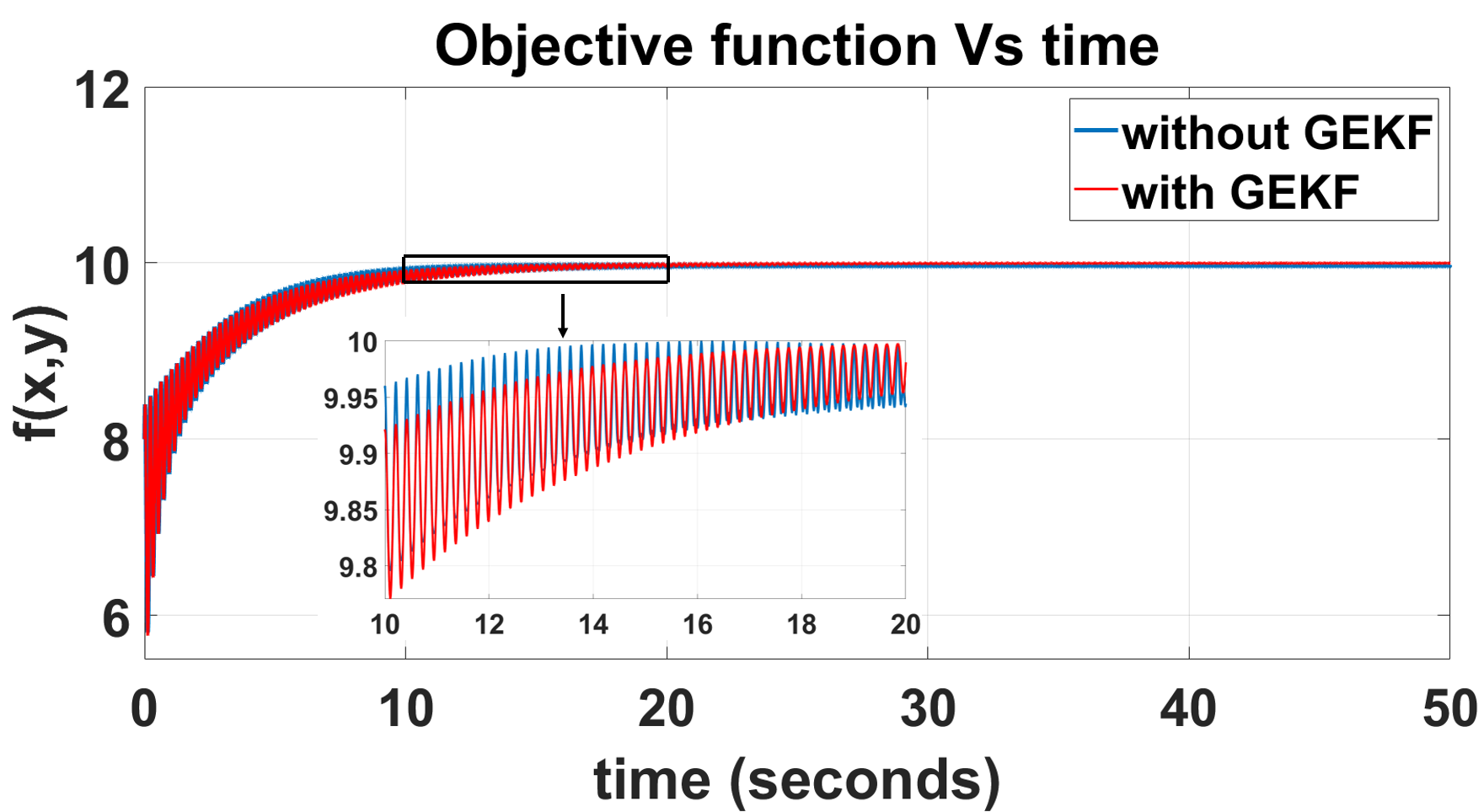}
  \end{subfigure}
  
  \begin{subfigure}{0.45\textwidth}
    \includegraphics[width=\textwidth]{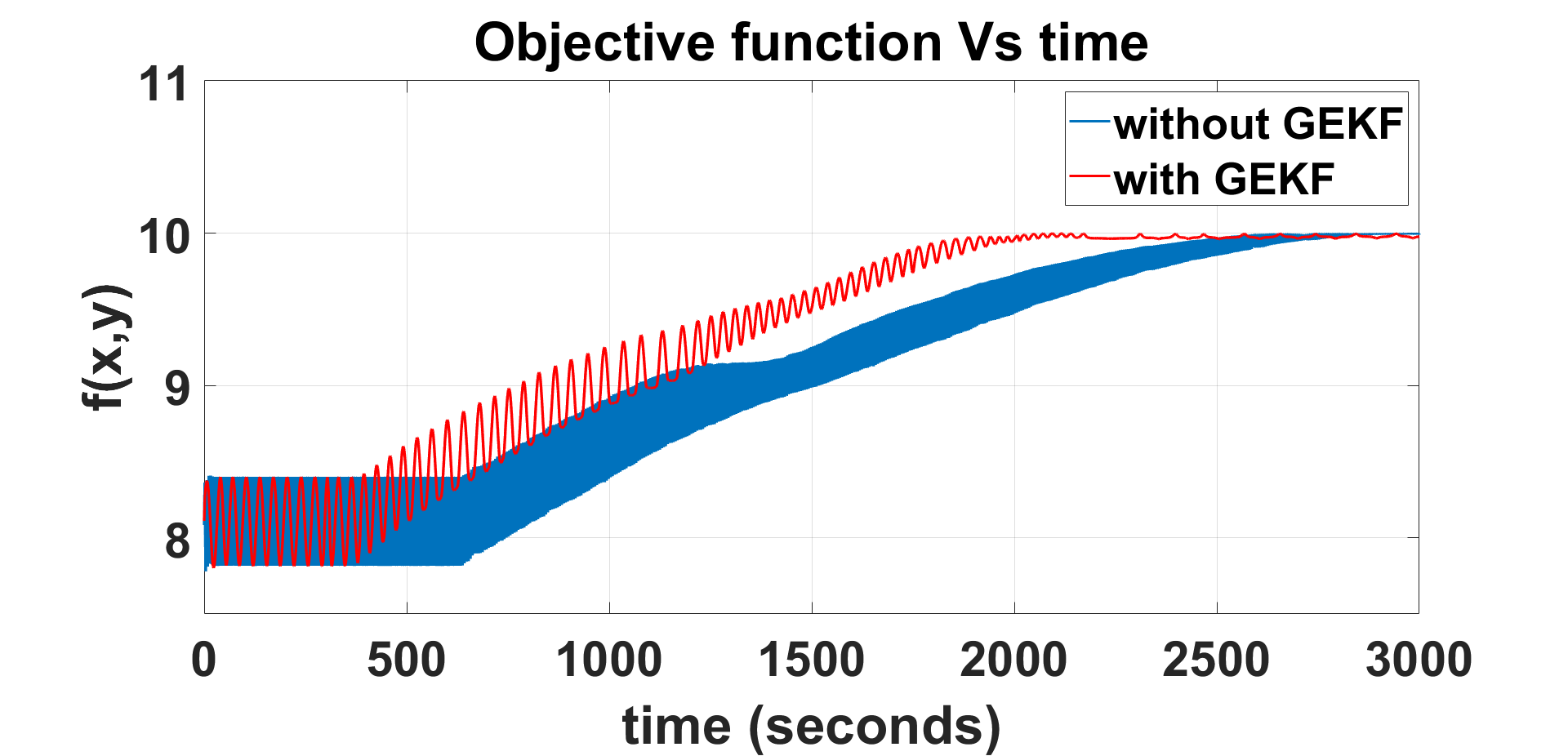}
  \end{subfigure}

  \caption{Objective Function vs. time (top for simulation, bottom for experimental implementation). The proposed design (red) is referred to by ``with GEKF" whereas the literature design is referred to by ``without GEKF."}
  \label{fig:Objective function MM_simulation and experiment}
 \end{figure}

The simulation results are provided in figure \ref{fig:Mathmatical_Model_simulation} parts (a)-(c) in addition to figure \ref{fig:Objective function MM_simulation and experiment} (top). 
 From figure \ref{fig:Mathmatical_Model_simulation}, it is obvious that our proposed design with GEKF attenuated the oscillations and asymptotically converged to the maximum points of $x$ and $y$. As expected, the $x$ vs. $y$ planner trajectory of the system shows convergence with attenuated oscillations. Lastly, the objective function reached its maximum $f^*=10$ with attenuated oscillations in figure \ref{fig:Objective function MM_simulation and experiment} (top). In all simulations, it is safe to conclude that our proposed design with GEKF outperformed the literature design (no GEKF) significantly in that it converged faster and with attenuated oscillations. 

\begin{figure*}[ht]
  \centering
  \begin{subfigure}{0.3\textwidth}
    \includegraphics[width=\linewidth]{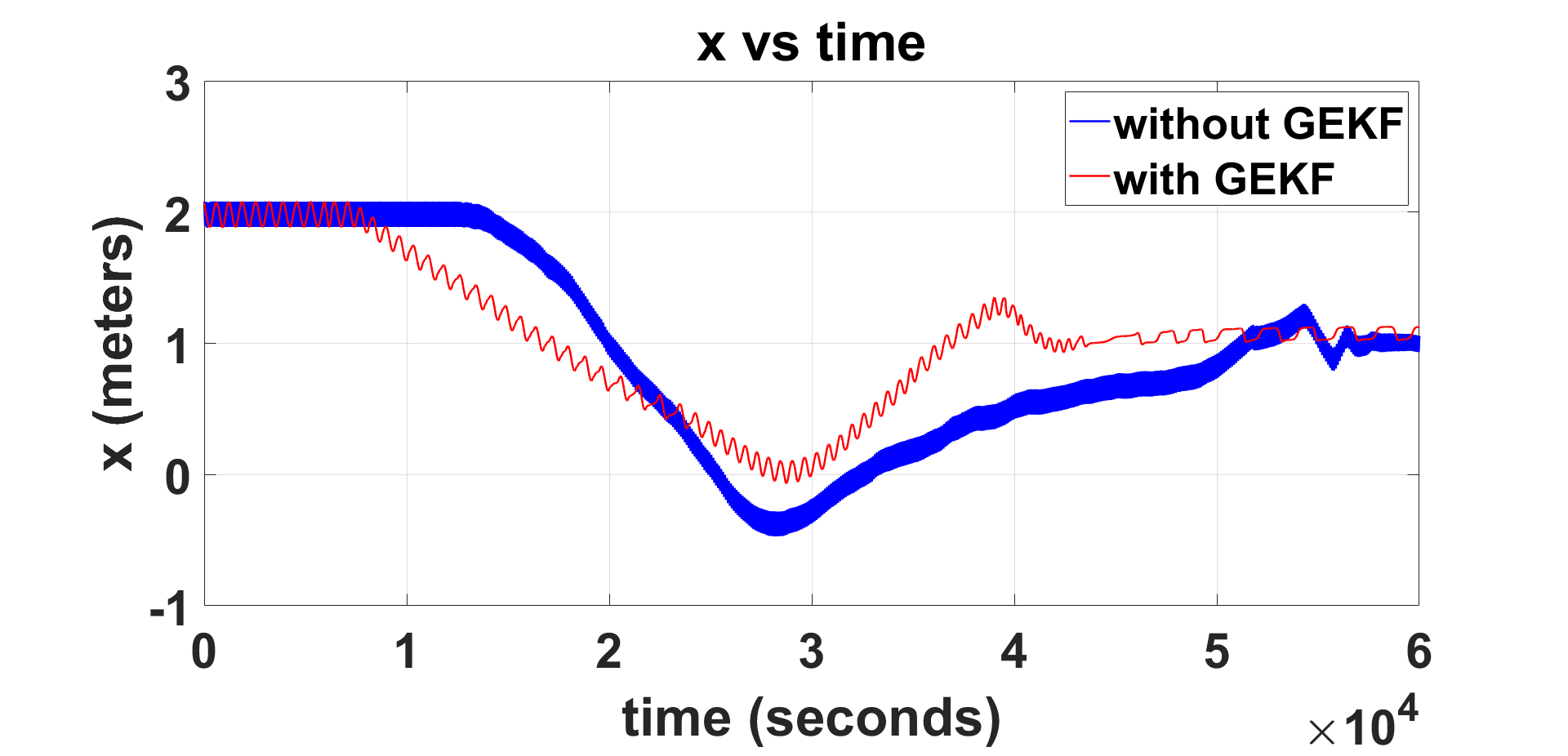}
    \caption{$x$ position vs. time.}
    \label{fig:X_Vehicle_2}
  \end{subfigure}
  \hfill
  \begin{subfigure}{0.3\textwidth}
    \includegraphics[width=\linewidth]{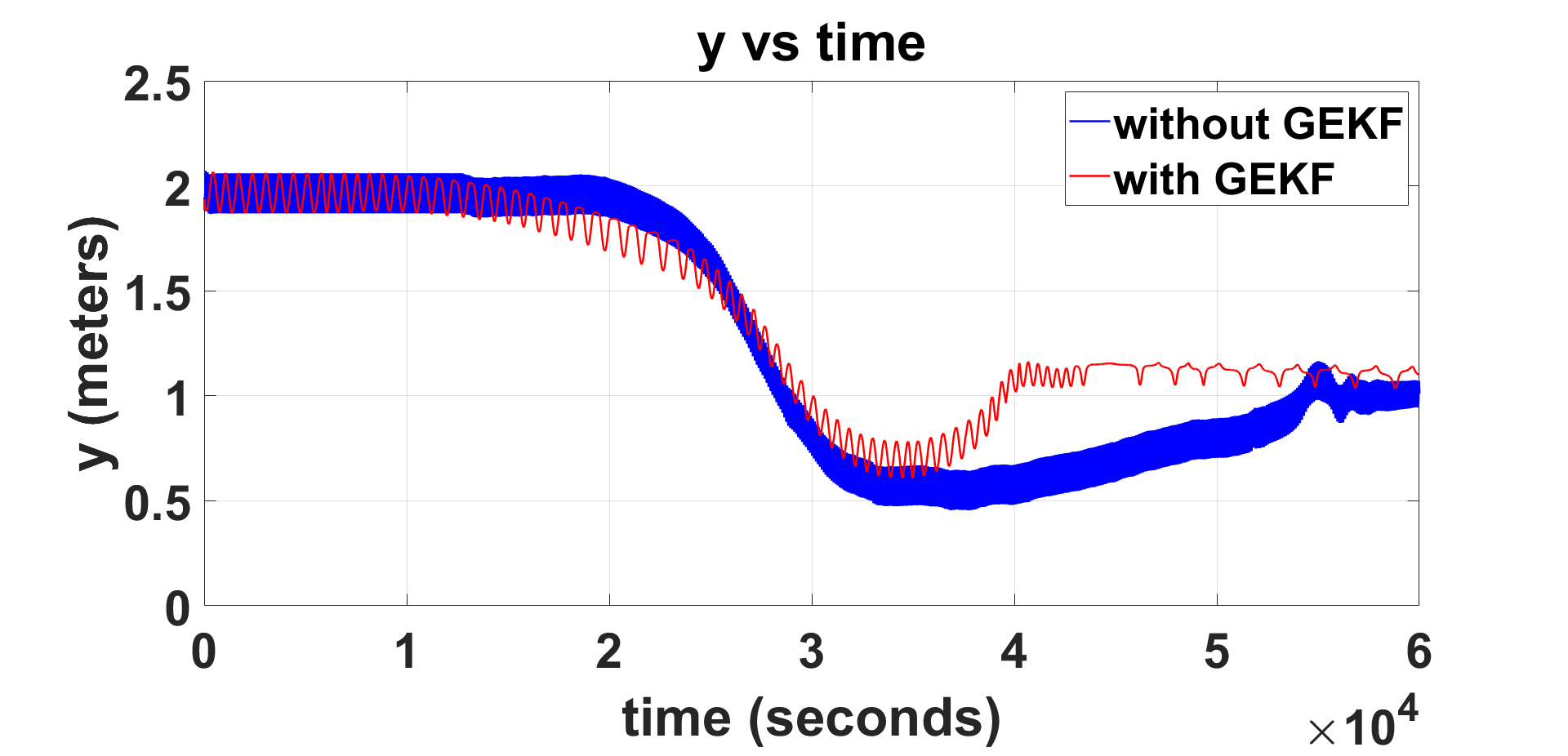}
    \caption{$y$ position vs. time.}
    \label{fig:Y_Vehicle_2}
  \end{subfigure}
  \hfill
  \begin{subfigure}{0.3\textwidth}
    \includegraphics[width=\linewidth]{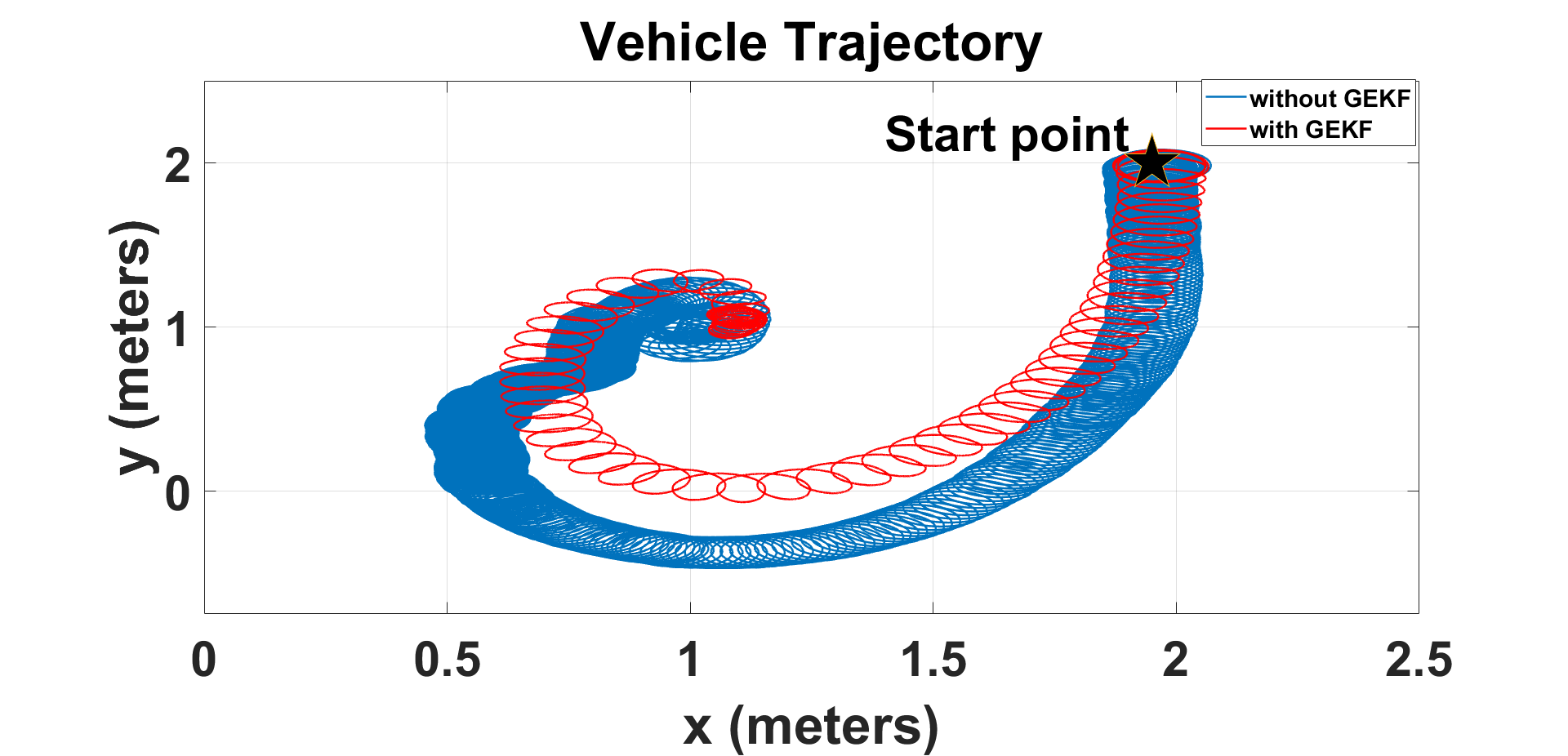}
    \caption{Planar plot of $x$ vs. $y$.}
    \label{fig:X_Y_V2}
     \end{subfigure}
 
  \caption{Trajectories (data obtained via motion capture system) of $x$ and $y$ coordinates of the real-world, real-time TB3 robotic experiment obtaining measurements from a known objective function. We did the experiment two times to compare our proposed design (red) with the literature design (blue). Clearly our design attenuate oscillations and converges faster.}
  \label{fig:Mathmatical_Model_Experiment}
\end{figure*}

Moving forward with the experimental part of this phase, we conducted real-world experiments in which we used the same known objective function that we used in the simulations for the mere purpose of obtaining measurements. In our experiments, we implemented the proposed single-integrator design with GEKF and compared its performance with the traditional single-integrator of the literature (no GEKF). For the experimental implementation, we used the following parameters. Initial values of $(x,y)$ states as (2,2), the covariance matrix $P$ as $[4,4,4,4,4]^T$, sample rate $T_{out} = 0.1$, $\bm{Q}=0.05\bm{I}$, $R=0.5$ and $N=10$. The ESC parameters are: $\omega=30$ and $c=0.5$. The adaptation law parameters: $a_x(0)=0.137$, $a_y(0) = 0.137$; note that for simulations with traditional single-integrator (i.e., no GEKF and no adaptation law), $a_x=constant=0.137$, $a_y=constant=0.137$, and the tuning parameter $\lambda_x=\lambda_y=0.005$. The two optional HPF were used such that $h_1=h_2=0.005$.

Our real-world, real-time experimental results are presented in figure \ref{fig:Mathmatical_Model_Experiment}. We can \textit{remarkably} observe the effectiveness of our proposed design for the first time experiments involving GEKF. It is clear that $x$ and $y$ states reached the desired maximum point with attenuated oscillations (i.e., achieving asymptotic convergence) as shown in parts (a) and (b) of the figure. Moreover, as shown in part (c) of the planar plot, the TB3 reached the maximum position taking a shorter and faster path (i.e., \textit{better convergence rate} compared to the literature design. Moreover, as shown in figure \ref{fig:Objective function MM_simulation and experiment} (bottom), the objective function could reach the optimal value faster and with \textit{better convergence rate} with our design when compared with the literature. Also, for the convenience of the reader and for better visualization, we made a YouTube video accessible in \cite{Video_mathmatical_model}. It is worth noting that the attenuation of oscillations and asymptotic convergence is the verification of the effectiveness of our design which is based on our theoretical findings in \cite{pokhrel2023control}. However, we here \textit{record} that ``\textit{better convergence rate}'' was not something that was studied before; this suggests that GEKF can have more and wider benefits to control-affine ESC systems. Hence, this paper suggests the need for future studies on the wider and broader effects of using GEKF in control-affine ESC systems.

\subsection{Experimentation of the Proposed Design for Model-free, Real-time Source Seeking of Light}

For more verification and validation of the effectiveness of our proposed design, we conducted another phase of experimentation. We conducted real-world, real-time experiments in which we used a light intensity/distribution as the unknown objective function which we only have access to its measurements via a light sensor; clearly, the extremum will be the maximum light intensity which will be the position of the source itself (a source seeking problem). That is, this experiment is totally model-free (no mathematical expression is used for the TB3, sensor, or light intensity/distribution). We implemented the proposed design with GEKF and compared it with another implementation using the literature design (no GEKF) for performance evaluation. During our experiments, we used the following parameters.
Initial values of $(x,y)$ states as (0,2), the covariance matrix $P$ as $[4,4,4,4,4]^T$, sample rate $T_{out} = 0.1$, $\bm{Q}=0.05\bm{I}$, $R=100$, and $N=10$. The ESC parameters are: $\omega=30$ and $c=1$. The adaptation law parameters are: $a_x(0)=0.027$, $a_y(0) = 0.027$, and $\lambda_x=\lambda_y=0.005$. The optional HPFs were used such that $h_1=h_2=1.5$. As done in the previous phase, for the experiment without GEKF, we took $a_x=constant=0.027$ and $a_y =constant= 0.027$.
\begin{figure*}[h]
  \centering
  \begin{subfigure}{0.32\textwidth}
    \includegraphics[width=\linewidth]{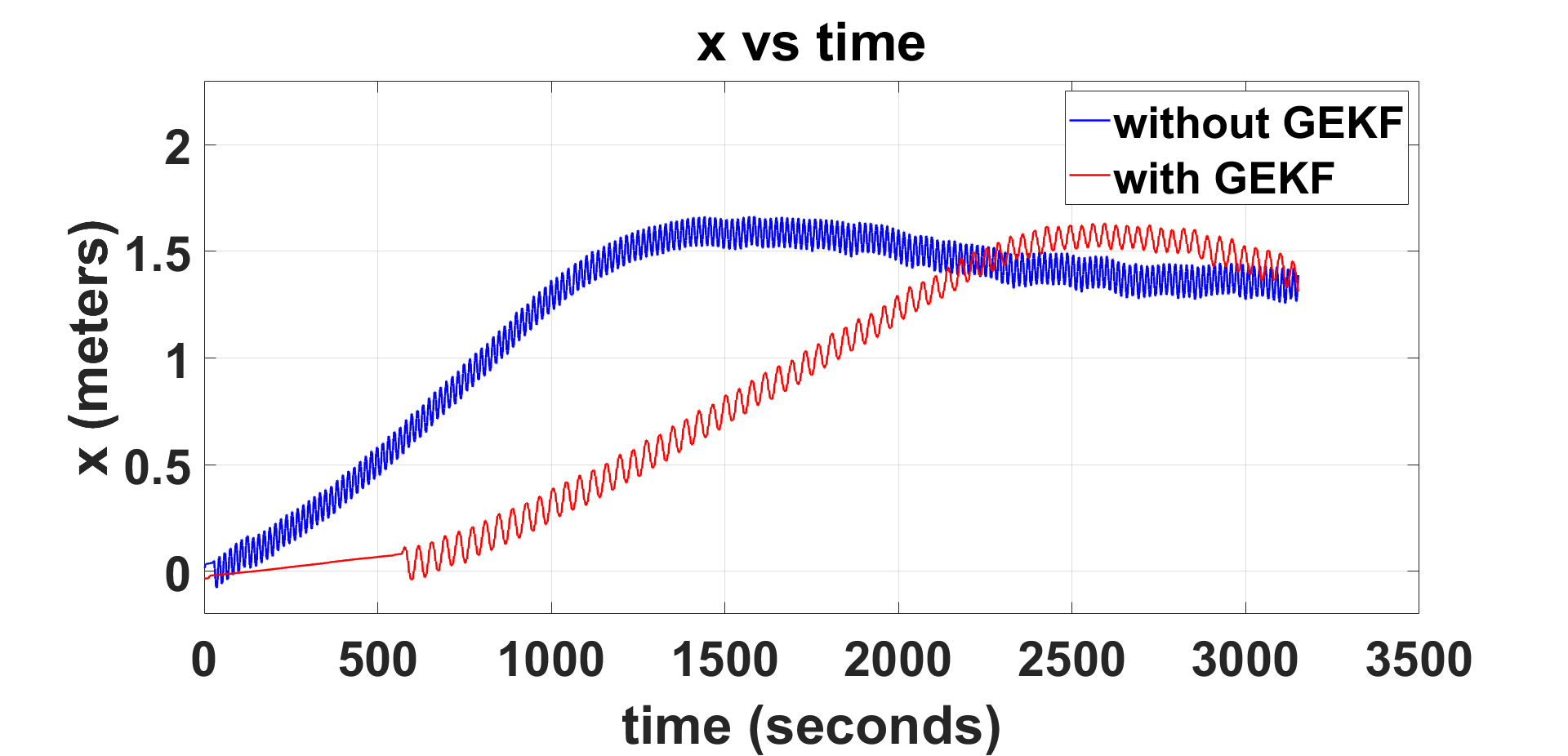}
    \caption{$x$ position vs. time.}
    \label{fig:x_position}
  \end{subfigure}
  \hfill
  \begin{subfigure}{0.32\textwidth}
    \includegraphics[width=\linewidth]{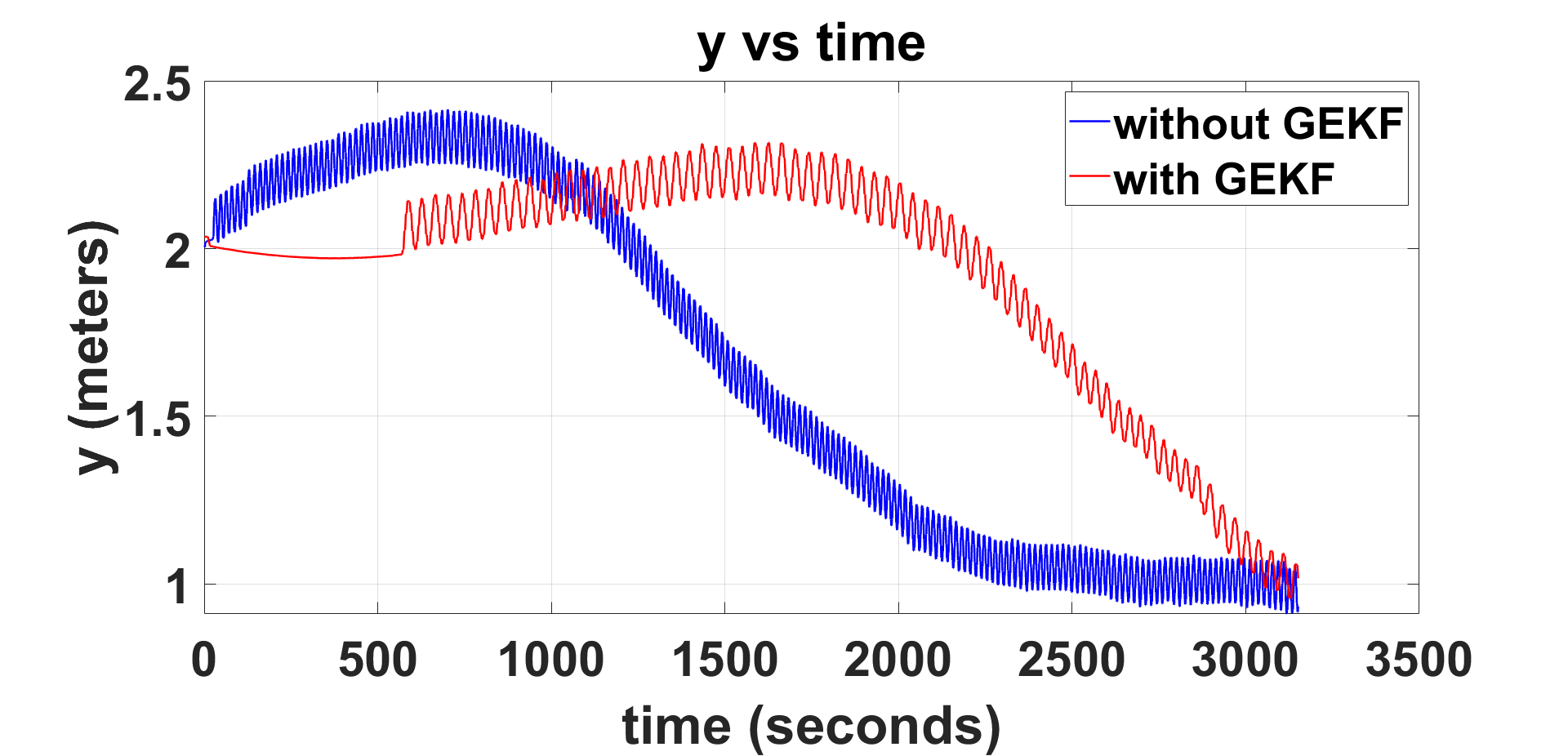}
    \caption{$y$ position vs. time.}
    \label{fig:Y_position}
  \end{subfigure}
  \hfill
  \begin{subfigure}{0.32\textwidth}
    \includegraphics[width=\linewidth]{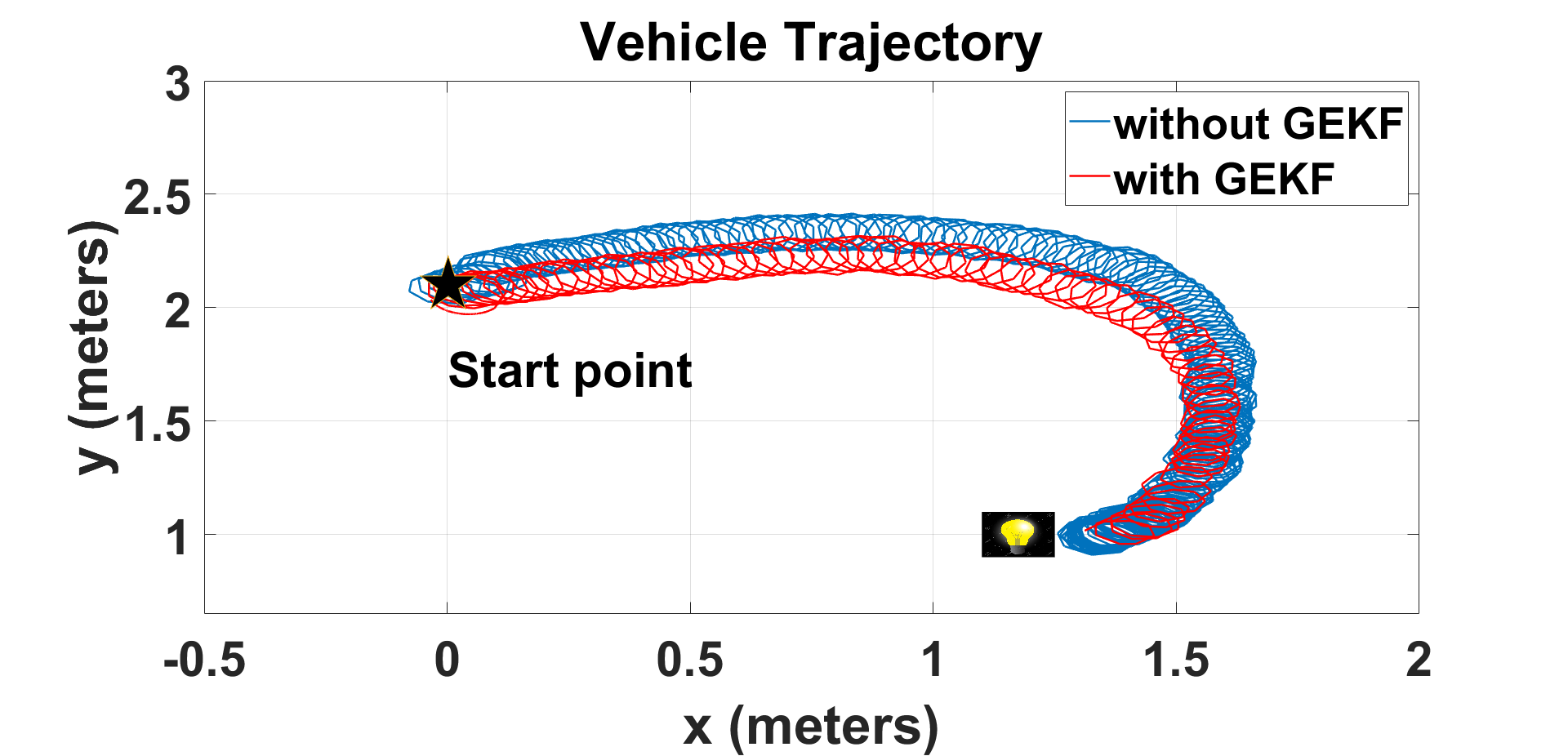}
    \caption{Planar plot of $x$ vs. $y$.}
    \label{fig:X_Y_V_9}
  \end{subfigure}
  
  \caption{Trajectories (data obtained via motion capture system) of $x$ and $y$ coordinates of the real-world, real-time TB3 robotic experiment obtaining measurements of the light intensity via sensor. We did the experiment two times to compare our proposed design (red) with the literature design (blue). Clearly our design attenuate oscillations and converges faster.}
  \label{fig:light source experiment}
\end{figure*}

\begin{figure*}[ht]
   \centering
  \begin{subfigure}{0.24\textwidth}
   \includegraphics[width=\textwidth]{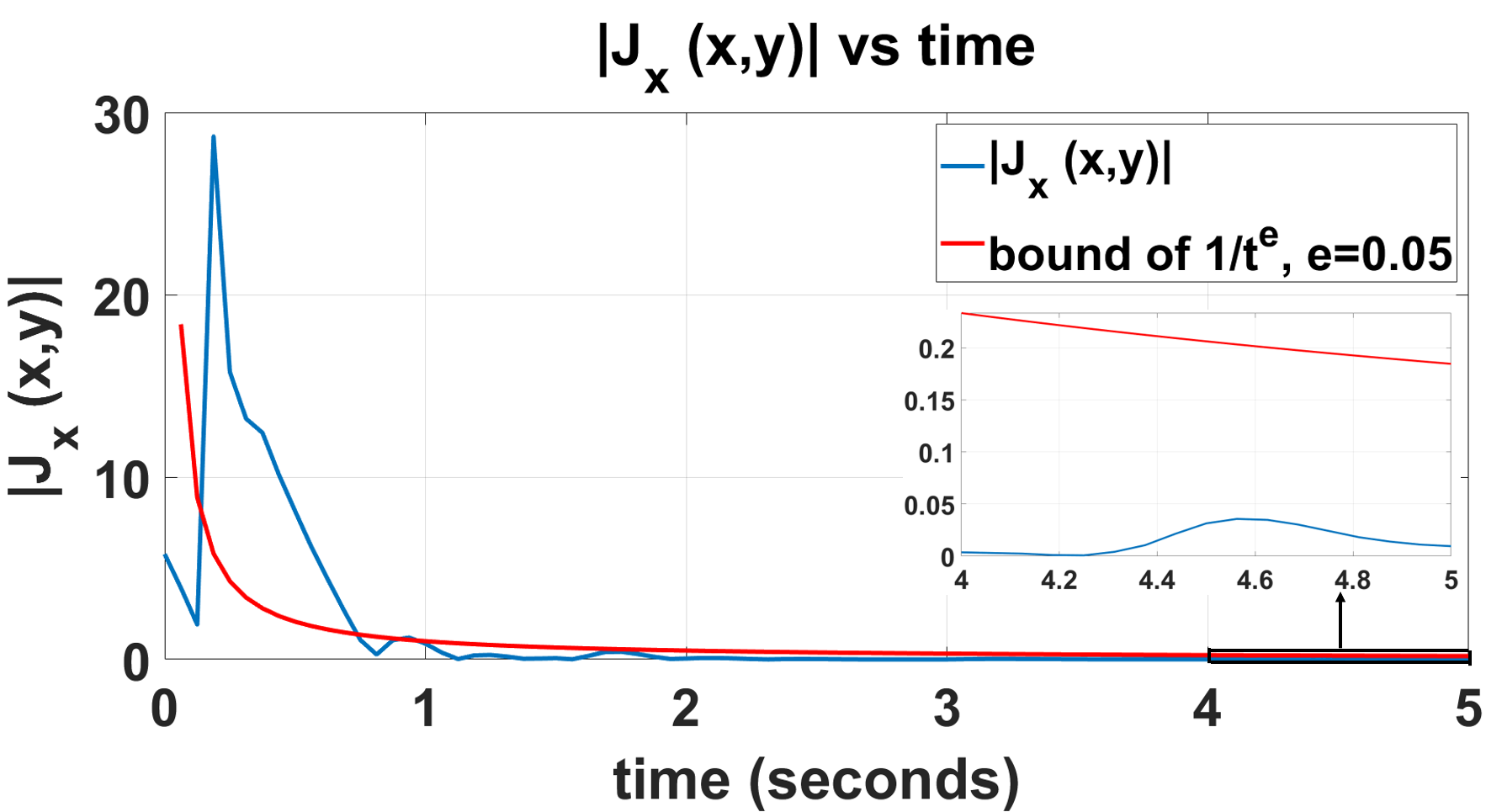}
    \caption{$J_x$ vs Time}
  \end{subfigure}
  \hfill
  \begin{subfigure}{0.24\textwidth}
    \includegraphics[width=\textwidth]{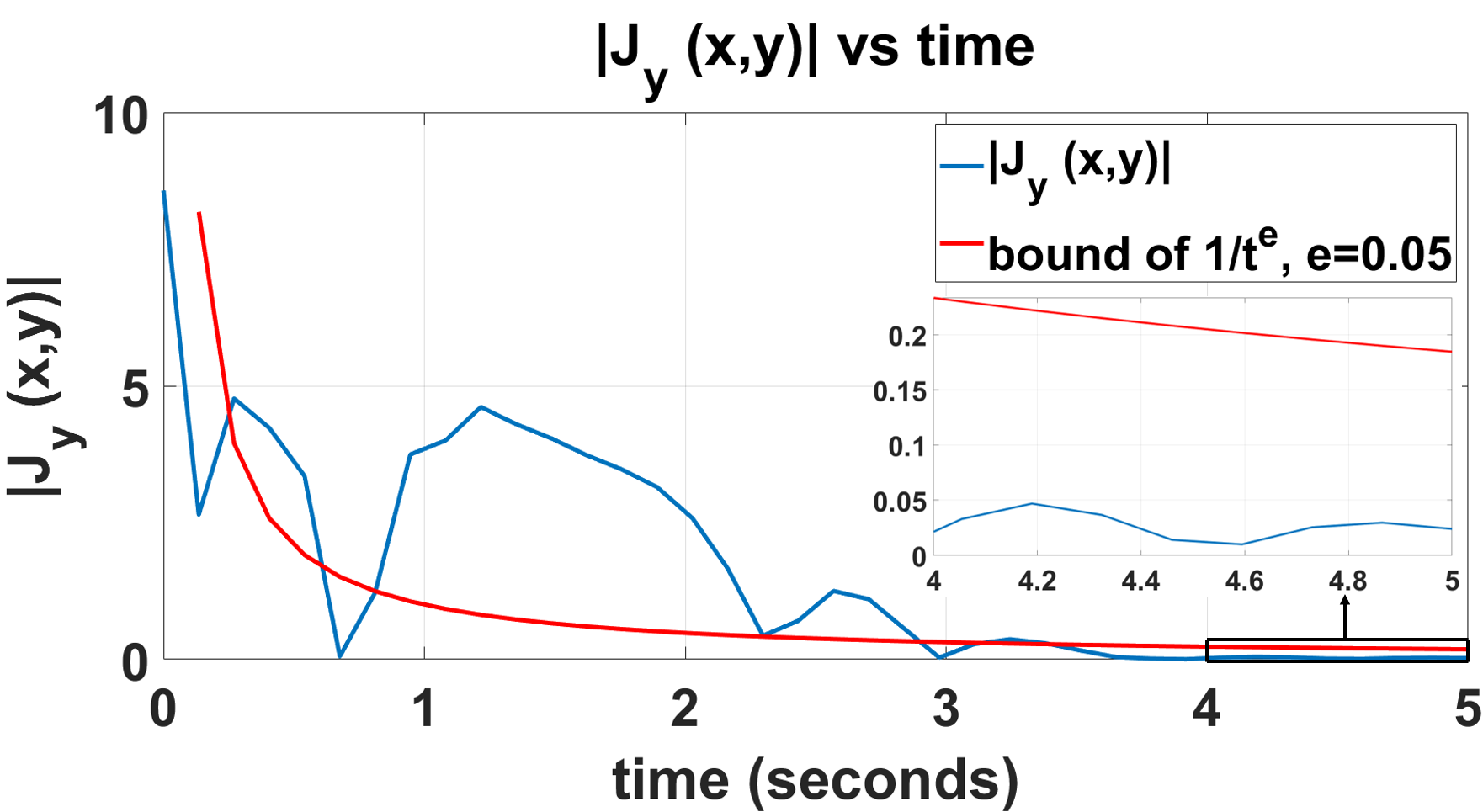}
    \caption{$J_y$ vs Time}
  \end{subfigure}
   \hfill
  \begin{subfigure}{0.24\textwidth}
    
     \includegraphics[width=\textwidth]{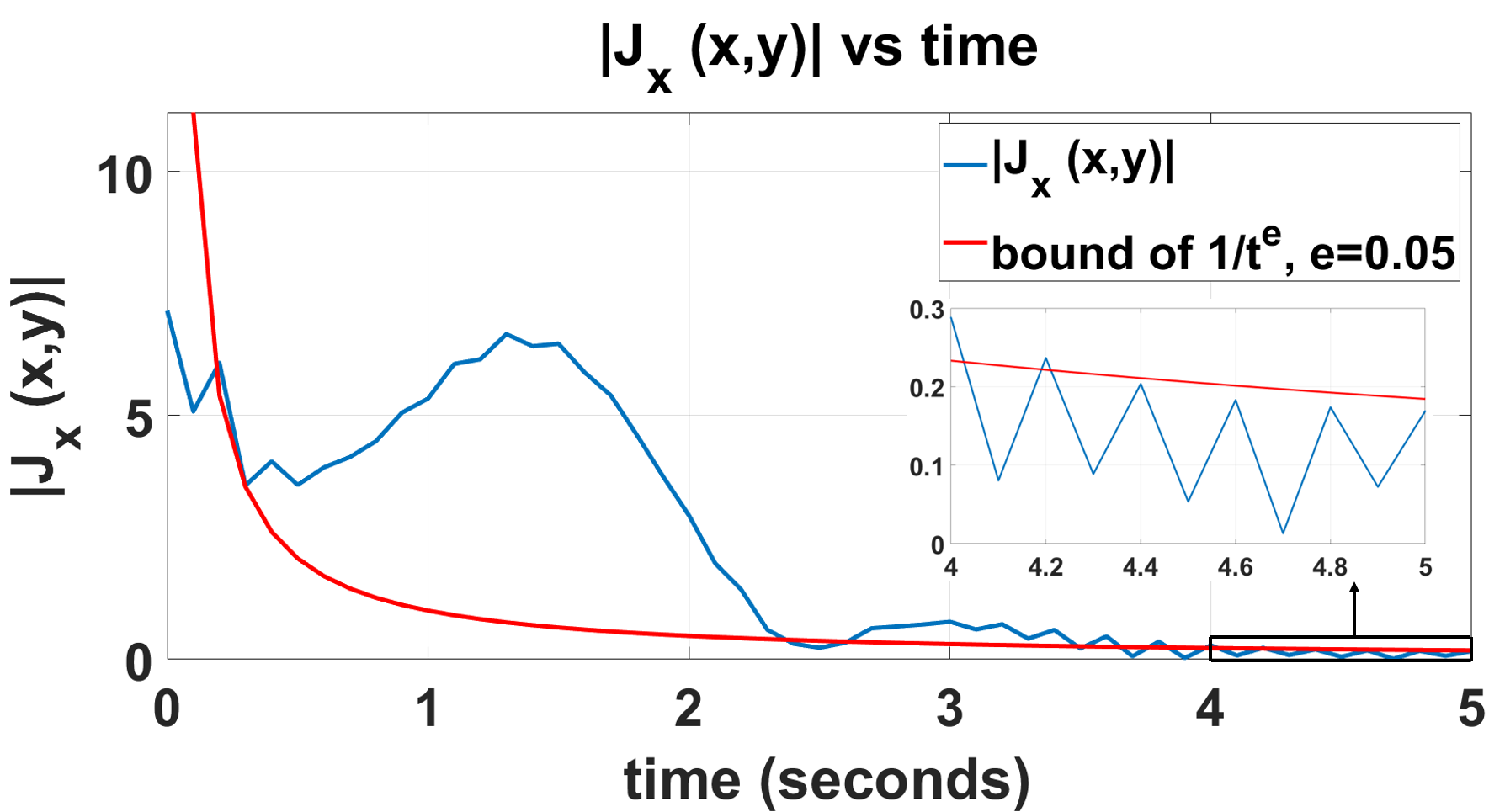}
    \caption{$J_x$ vs Time}
  \end{subfigure}
  \hfill
  \begin{subfigure}{0.24\textwidth}
  \includegraphics[width=\textwidth]{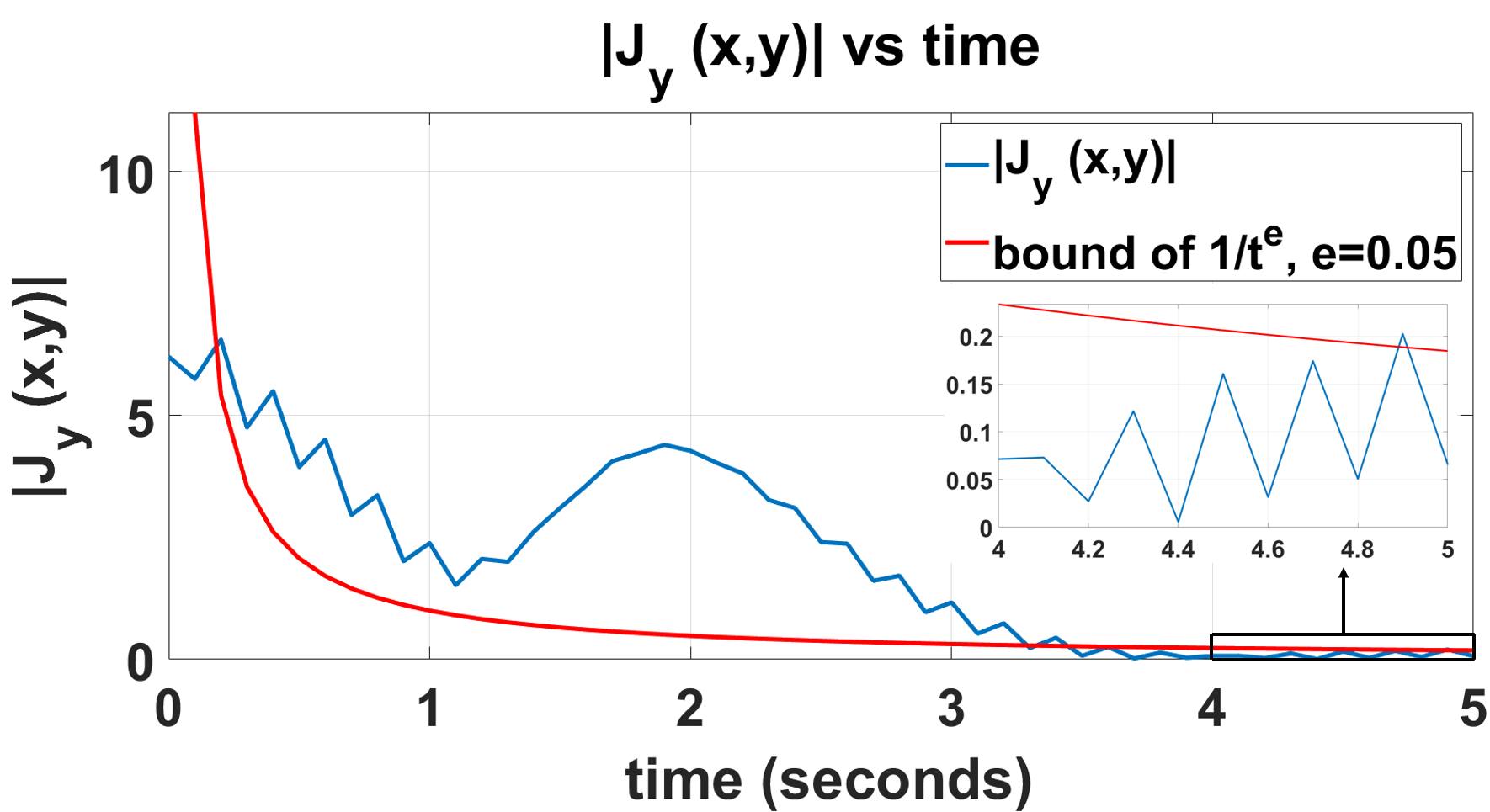}
    \caption{$J_y$ vs Time}
  \end{subfigure}
  \hfill
  \caption{$J_x, J_y$ vs. against bound $\frac{1}{t^e}$ for known objective function experiment (a,b) and light source experiment (c,d).}
  \label{fig:stability_mathmatical model and light_source_experiment}
 \end{figure*}
Our novel real-world, model-free, real-time experiments showed \textit{remarkably} promising results in two fronts. First, our proposed design could steer the TB3 to the light source with attenuated oscillations (i.e., with asymptotic convergence) with better performance in $x$ position as shown in part (a) of figure \ref{fig:light source experiment} and in $y$ as shown in part (b) of figure \ref{fig:light source experiment}; this is also clear in the planner plot in part (c) of the same figure. Second, which is, again, an observation we made in the previous phase (subsection 3.B), the TB3 reached the maximum position (the light source itself) taking a shorter and faster path which demonstrates our GEKF design has better \textit{convergence rate}. It is noticeable, however, that due to some software and/or hardware issues/limitations, a slight delay occurred while using the GEKF as shown in figure \ref{fig:light source experiment} (see the red graph in parts a and b). Despite this early delay, the GEKF subsequently improved the convergence rate and successfully attenuated the oscillations. It is significant to record here that our proposed design maintained the same observation we made in the previous phase about ``better convergence rate" even though, clearly, source seeking of light via a sensor is less ideal of a condition to experimentation, due to, for example, more noises and sensitivities to the environment. For the convenience of the reader and better visualization, we made a YouTube video for this experiment which is accessible in \cite{Video_light_source}, the first link.

\begin{figure}[ht]
    \centering
    \includegraphics[width=0.45\textwidth]{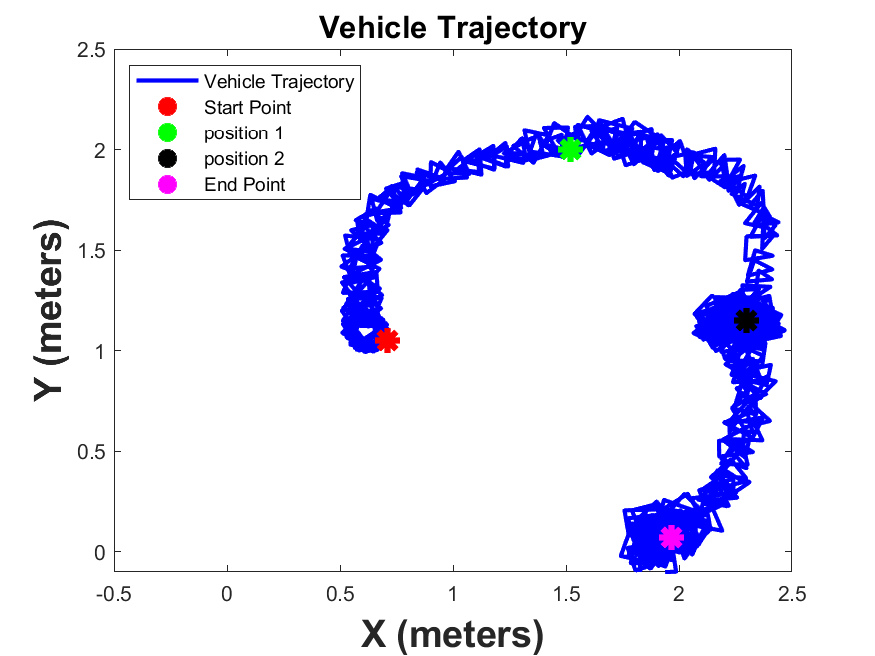}
    \caption{Planar plot for light source seeking by our proposed design when the source keep changing its position with time. The data captured for the trajectories via the motion capturing system are presented here for the deployed TB3.}
    \label{fig:multiple position} 
\end{figure}

We also conducted another experimentation to further test the capability of the proposed design. It is clear that as the TB3 reaches the light source (the extremum), it will have attenuated oscillations. But what if the light source (the extremum) moves with time (i.e., time varying extremum)? Our adaptation law (see figure \ref{fig:ESC Design with EGKF} and Eq \eqref{eqn:eg3esc}) changes the rate of the amplitude of the input signal depending on how far the system is from the extremum (i.e., the closer the system is from the extremum, the smaller oscillations it needs to extract gradient information). However, from Eq \eqref{eqn:eg3esc}, it is clear that the vice-versa situation may be applicable as well. That is, if we start from a small values of $a_x$ and $a_y$ and we are far away from the extremum, the adaptation law will increase the rate of $a_x$ and $a_y$ to increase the oscillations so that more gradient information can be extracted. Our experiment succeeded in demonstrating the capability of our design to do so and to track the source by increasing or decreasing the oscillations as needed. This suggests that future studies can be done to generalize further our theoritical results in \cite{pokhrel2023control} to admit time varying extremum. Our results are provided in figure \ref{fig:multiple position}. Also, we made a YouTube video for better visualization that can be accessed in \cite{Video_light_source}, the second link.

\subsection{Comments on Stability from Experimentation}

In literature, the stability of single-integrator without GEKF has been studied/concluded in \cite{durr2013lie} and was generalized in \cite{grushkovskaya2018class}. However, for the proposed design with GEKF which is based on the theoretical results provided in our previous work \cite{pokhrel2023control}, Theorem 1 can be used for verification of stability. Theorem 1 necessitates that the terms $|J_x(x,y)|$ and $|J_y(x,y)|$ be bounded eventually after some time $t^*$ by a decreasing function $1/t^P$ with $P>1$. Thus, we plotted said components from two experiments: the one from subsection 3.B using the measurements of a known objective function (its results are provided in figure \ref{fig:Mathmatical_Model_Experiment}); and the one from subsection 3.C using light intensity/distribution measurements via a sensor (its results are provided in figure \ref{fig:X_Y_V_9}). The plots show that the components $|J_x|$ and $|J_y|$ are behaving as expected per Theorem 1 -- see figure  \ref{fig:stability_mathmatical model and light_source_experiment}. Parts (a) and (b) are from the first experiment (figure \ref{fig:Mathmatical_Model_Experiment}), whereas parts (c) and (d) are from the second experiment (figure \ref{fig:X_Y_V_9}).

\section{Conclusive Remarks and Future Work}
This paper provided a novel design with experimental verification for model free, real-time source seeking via control-affine ESC systems. The paper provided real-world experiments by a robot (TB3) for said design. Experimentation of the proposed design has shown promising results verifying the effectiveness of our recent theoretical results in \cite{pokhrel2023control} which allows for attenuating the oscillation of the ESC system based on gradient estimation via what we call Geometric-based Kalman Filter (GEKF) \cite{pokhrel2023gradient}. To the best of our knowledge, this paper is the first to provide an experimental validation that single-integrator-like control-affine ESC system can perform well. In fact, it was argued and shown in \cite{grushkovskaya2018family} that the simple control law used for example in single-integrator designs ($\dot{x}=u=$$f(x)\sqrt{\omega} u_1+\sqrt{\omega} u_2$) are not recommended. This has also been observed in our experiments when we implemented traditional single-integrator of the literature. However, we observed and recorded \textit{remarkable and significant improvement} in performance using our proposed design, which is an amended single-integrator structure with GEKF. Our results concluded that, not only the attenuation of oscillations have been achieved successfully, but also, the proposed design possesses \textit{better convergence rate}.

In the future, and encouraged by the results of this paper, multiple studies can take place. First, theoretical analysis should be conducted to provide the necessary conditions and proof that the use of GEKF with control-affine systems improve the convergence rate. Second, given how successful and effective the use of GEKF have been with simple design and simple control laws of single-integrator, it is recommended justifiably to try amending other control-affine ESC designs using GEKF in a similar manner to this paper. Lastly, it is important to develop a theoretical, and perhaps, experimental works studying the necessary conditions for admitting time-varying objective functions with time-varying extremum to control-affine ESC systems with GEKF (i.e., further generalization of our work in \cite{pokhrel2023control}).   


\bibliography{references}
\bibliographystyle{IEEEtran}
\end{document}